\documentclass[10pt,twocolumn,letterpaper]{article}

\usepackage{cvpr}              % To produce the CAMERA-READY version
\definecolor{cvprblue}{rgb}{0.21,0.49,0.74}
\usepackage[pagebackref,breaklinks,colorlinks,allcolors=cvprblue]{hyperref}
\usepackage{graphicx}
\usepackage{caption}
\usepackage{lipsum}
\usepackage{amsthm}
\usepackage{amssymb}
\usepackage{algorithm}
\usepackage{algorithmic}
\usepackage[algo2e, boxed, ruled]{algorithm2e}
\usepackage{booktabs}
\usepackage{amsmath}
\usepackage{multirow} 
\def\paperID{****} % *** Enter the Paper ID here
\def\confName{CVPR}
\def\confYear{2025}
\usepackage{ulem}
\title{VideoMaker: Zero-shot Customized Video Generation with the Inherent Force of Video Diffusion Models}
\author {
    Tao Wu \textsuperscript{\rm 1,2 *},
    Yong Zhang \textsuperscript{\rm 3 *},
    Xiaodong Cun \textsuperscript{\rm 3 *},
    Zhongang Qi \textsuperscript{\rm 4 $\dag$},
    Junfu Pu \textsuperscript{\rm 2},
    \\
    Huanzhang Dou \textsuperscript{\rm 1},
    Guangcong Zheng \textsuperscript{\rm 1},
    Ying Shan\textsuperscript{\rm 2,3},
    Xi Li \textsuperscript{\rm 1 $\dag$} 
    \\
    $^1$College of Computer Science and Technology, Zhejiang University \\
    $^2$ARC Lab, Tencent PCG \quad
    $^3$Tencent AI Lab \quad
    $^4$Huawei Noah’s Ark Lab
}

\begin{document}
\twocolumn[{
\renewcommand\twocolumn[1][]{#1}
\maketitle
\begin{center}
    \centering
    \vspace*{-.8cm}
    \includegraphics[width=0.96\textwidth]{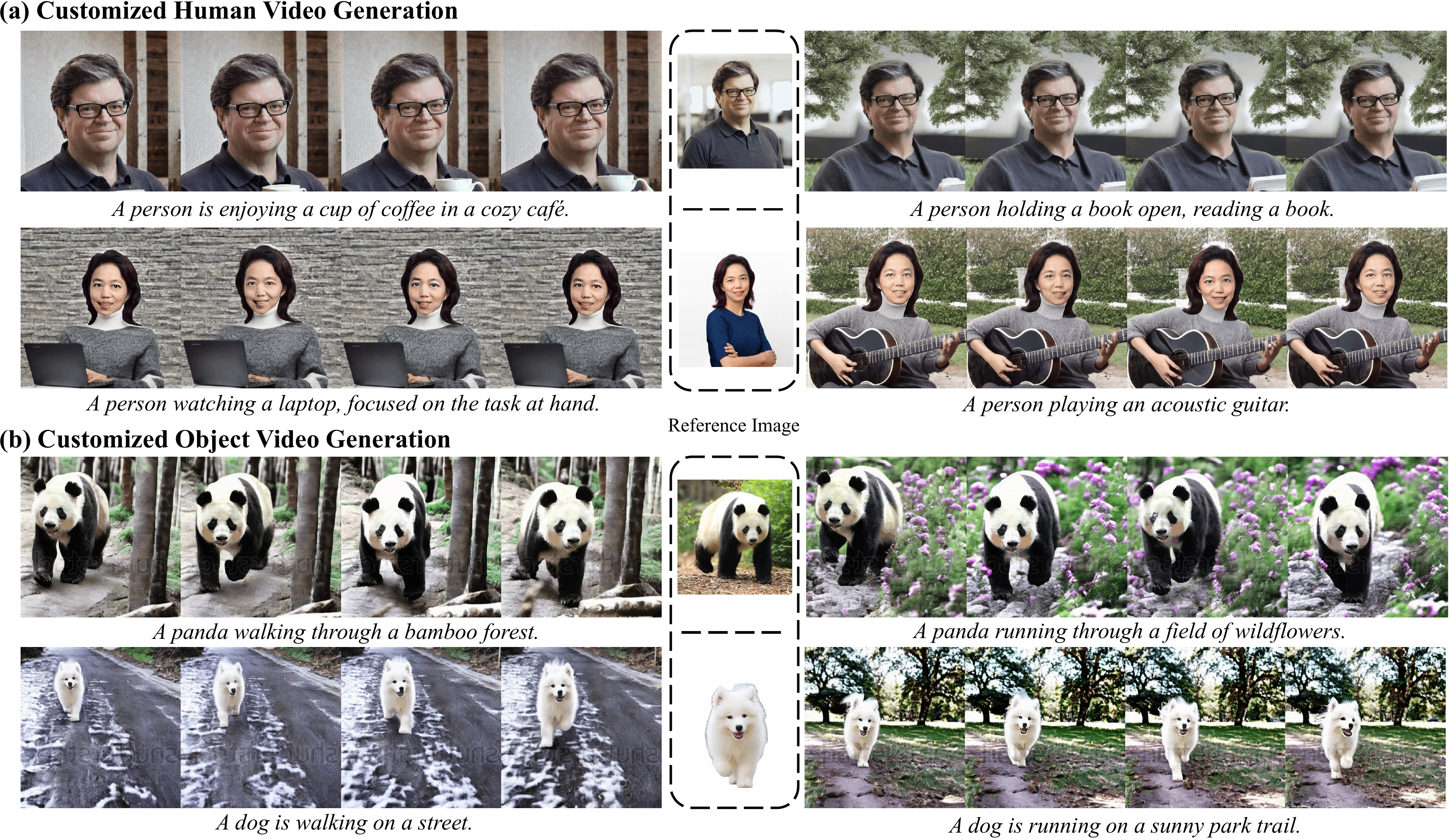}
    \vspace*{-.3cm}
    \captionof{figure}{Visualization for our VideoMaker. Our method achieves high-fidelity zero-shot customized human and object video generation based on AnimateDiff~\cite{hu2024animate}.}
\label{fig:teaser}
\end{center}
}]
\footnotetext[1]{$^*$ These authors contributed equally. $^\dagger$ Corresponding author.}

\footnotetext[2]{Work done during Zhongang Qi's tenure at Tencent PCG ARC Lab.}

% \newcommand{\thefootnote}{\arabic{footnote}}
% 我们的方法在AnimateDiff上实现了高保真的定制化视频生成。
\begin{abstract}

Zero-shot customized video generation has gained significant attention due to its substantial application potential. 
Existing methods rely on additional models to extract and inject reference subject features, assuming that the Video Diffusion Model (VDM) alone is insufficient for zero-shot customized video generation. 
However, these methods often struggle to maintain consistent subject appearance due to suboptimal feature extraction and injection techniques. 
In this paper, we reveal that VDM inherently possesses the force to extract and inject subject features. 
Departing from previous heuristic approaches, we introduce a novel framework that leverages VDM’s inherent force to enable high-quality zero-shot customized video generation. 
Specifically, for feature extraction, we directly input reference images into VDM and use its intrinsic feature extraction process, which not only provides fine-grained features but also significantly aligns with VDM’s pre-trained knowledge. 
For feature injection, we devise an innovative bidirectional interaction between subject features and generated content through spatial self-attention within VDM, ensuring that VDM has better subject fidelity while maintaining the diversity of the generated video.
Experiments on both customized human and object video generation validate the effectiveness of our framework.

\end{abstract}
\vspace{-0.4cm}

\section{Introduction}
\label{sec:intro}

Video Diffusion Models (VDMs)~\cite{guo2023animatediff,chen2024videocrafter2,wang2023modelscope,videoworldsimulators2024,yang2024cogvideox} can generate high-quality videos from a given text prompt. 
However, these pretrained models unable to create specific videos from a given subject since this customized subject is hard to be described by a text prompt only. 
This problem is so-called customized generation and has been explored by personalized fine-tuning~\cite{chefer2024still,wei2023dreamvideo,wu2024customcrafter,ruiz2023dreambooth}. 
Yet, the time-consuming subject-specific finetune limits its usage in the real world.
Recently, Some methods~\cite{jiang2024videobooth,he2024id} based on~\cite{ye2023ip,wang2024instantid} have initially explored zero-shot customized video generation. 
But these methods still fail to maintain a consistent appearance with the reference subject.

\begin{figure}[tb]
    \centering
    \includegraphics[width=1.0\linewidth]{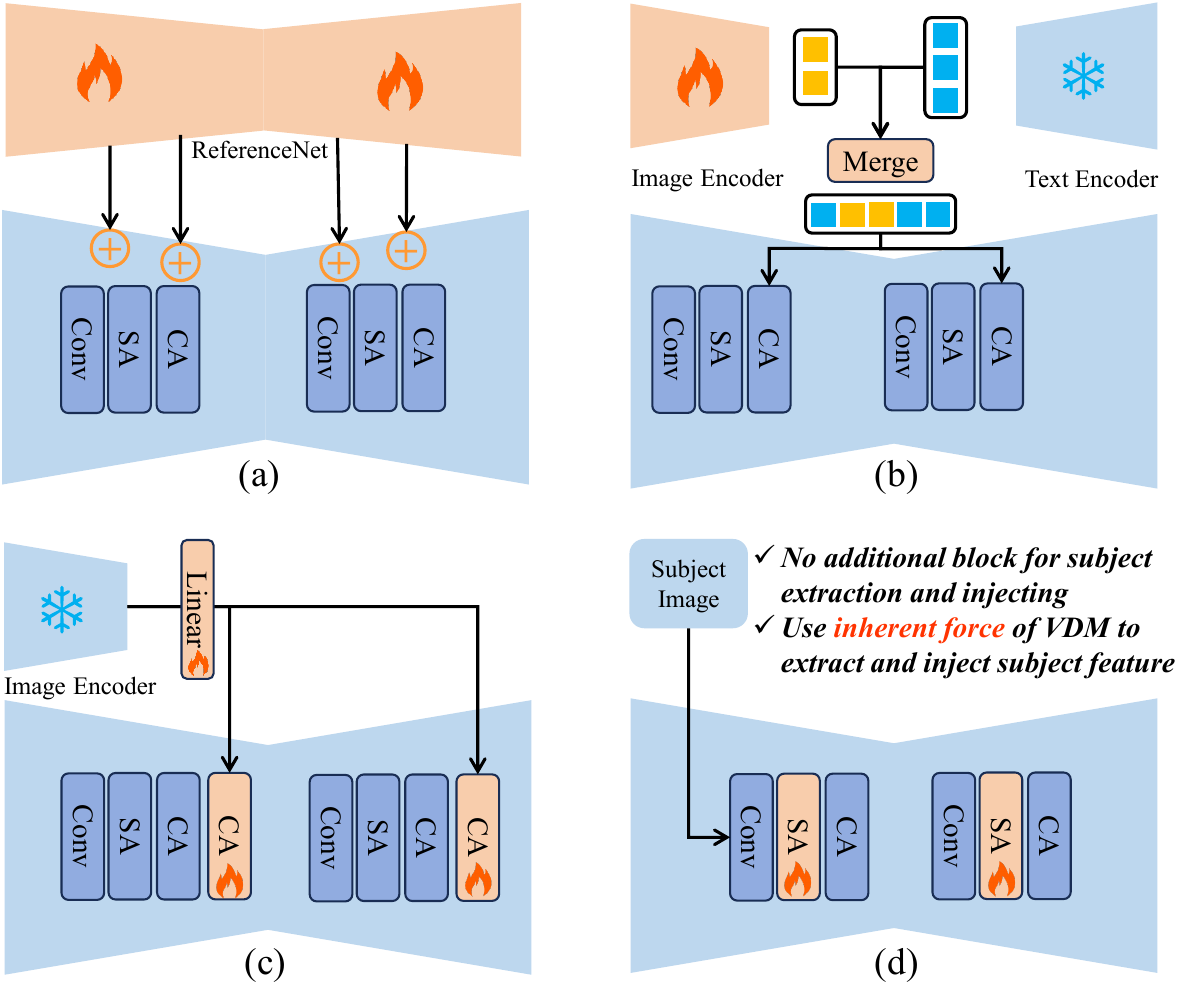}
    \caption{Compared with the existing zero-shot customized generation framework. Our framework does not require any additional modules to extract or inject subject features. It only needs simple concatenation of the reference image and generated video, and VDM's inherent force is used to generate custom video.}
    \label{fig:intro}
    \vspace{-0.2cm}
\end{figure}

Two keys for customized video generation are \textbf{subject feature extraction} and \textbf{subject feature injection}.
% Current methods assume that VDMs lack these capabilities, thus they heuristically rely on additional models to extract and inject subject features.
Current methods rely on additional models to extract and inject subject features, often overlooking the inherent capabilities of VDMs.
% Current approaches heuristically rely on additional models to extract and inject topic features.
\eg, some methods~\cite{hu2024animate,xie2024x,wang2024instantid} inspired by~\cite{zhang2023adding}, employ an additional ReferenceNet for feature extraction and directly add the subject features to the VDMs for injection~(\Cref{fig:intro}~(a)).
However, these methods introduce numerous additional training parameters, and this pixel-wise injection method significantly restricts the diversity of the generated videos.
Other methods~\cite{ye2023ip,li2024photomaker,jiang2024videobooth,he2024id} employ the pre-trained cross-modal alignment model~\cite{radford2021learning,xu2023metaclip,ma2024mode,ma2024ovmr} as feature extractors and inject subject feature by cross-attention layer (\Cref{fig:intro}~(b,c)).
Nevertheless, these methods produce only coarse-grained, semantic-level features from the pre-trained extractor, which fail to capture the details of the subject.
Consequently, these well-designed heuristic methods have not achieved satisfactory results in customized video generation. 
A question naturally arises: \textit{Perhaps VDMs have the force to extract and inject subject features, and we only need to activate and use these forces in a simple way to achieve customized generation ?}

Rethinking the VDMs, we identified some potential inherent forces.
% Rethinking the VDMs, we identified certain inherent force that can be harnessed.
For subject feature extraction, since inputting a noise-free reference image can be seen as a special case with a timestep of 0, the pre-trained VDM is already capable of extracting features from this without additional training. 
For subject feature injection, the spatial self-attention in the VDM primarily models the relationships between different pixels within a frame, making it more suitable for injecting subject reference features that are closely related to the generated content. 
Moreover, due to the self-adaptive nature of spatial self-attention, it can selectively interact with these features, which helps prevent overfitting and promotes diversity in the generated videos.
Therefore, if we utilize VDM itself as a fine-grained feature extractor for the subject and then interact the subject features with the generated content through spatial self-attention, we can leverage the inherent force of VDM to achieve customized generation.
% 因此，一如
% Therefore, our task is to design a simple and reasonable framework that cleverly leverages these inherent force of VDMs to achieve customized video generation.

% for rebuttall； cross-attention在原生VDM中是处理跨模态信息的，如果要用cross-attention还得新增，有现成的干啥不用。
Inspired by the above motivation, we present our VideoMaker, a novel framework that leverages the inherent force of VDMs to enable high-quality zero-shot customized generation.
We register the reference image as part of the model's input, utilizing the VDM itself for feature extraction. 
The extracted features are not only fine-grained but also closely aligned with the VDM’s inherent knowledge, eliminating the need for additional alignment.
For subject feature injection, we use VDM's spatial self-attention to explicitly interact with the subject feature close to VDM’s inherent knowledge with the generated content when generating content per frame.
Additionally, to ensure the model can effectively distinguish between reference information and generated content during training, we have designed a simple learning strategy to enhance performance further.
Our framework employs a native approach to complete subject feature extraction and injection without adding additional modules. It only requires fine-tuning the pre-trained VDM to activate the model's inherent force.
Through extensive experiments, we provide both qualitative and quantitative results that demonstrate the superiority of our method in zero-shot customized video generation.
Our contributions are summarized as follows:
\begin{itemize}
    % 我们使用the inherent force of VDM完成subject细粒度外貌特征提取,切所提取的subject外表信息对VDM更友好。
    % 我们颠覆了之前依赖Cross-Atten和特征相加的方式进行信息注入的方式,创新性的使用VDM中原生的self-attention的计算机理进行subject完成subject信息注入。
    % 我们的方法无需额外引入任何模块或预训练权重进行subject提取或信息注入，仅依靠微调部分参数即可达成高质量的定制化视频生成。
    % 我们的方法
    \item We use the inherent force of the video diffusion model to extract fine-grained appearance features of the subject, and the extracted subject appearance information is more friendly to learn for the video diffusion model.
    \item We revolutionize the previous method of information injection, innovatively using the native spatial self-attention computation mechanism in the video diffusion model to complete subject feature injection.
    \item Our framework outperforms existing methods and achieves high-quality zero-shot customized video generation by only fine-tuning some parameters.
    % \item  Our framework does not require the introduction of any additional modules or pre-trained weights for subject features extraction or injection. High-quality customized video generation can be achieved simply by fine-tuning some parameters.
    % 大量的实验证明，我们的framework性能超越现有的zero-shot 定制化视频方法,依靠微调部分参数即可达成高质量的定制化视频生成。
    
\end{itemize}

\section{Related Work}
\subsection{Text-to-video diffusion models}
With the progress in diffusion models and image generation ~\cite{rombach2022high, ramesh2022hierarchical, podell2023sdxl, mou2024t2i, peebles2023scalable, nichol2022glide, wu2024spherediffusion,huang2024groupdiffusiontransformersunsupervised,huang2024incontextloradiffusiontransformers,dou2024gvdiffgroundedtexttovideogeneration,yuan2024semanticmimmarringmaskedimage,zhao2024towards}, there have been significant advancements in text-to-video (T2V) generation. 
Given the limited availability of high-quality video-text datasets~\cite{blattmann2023align,ju2024miradata}, numerous researchers have tried to develop T2V models by leveraging existing text-to-image (T2I) generation frameworks. 
Some studies~\cite{zhou2022magicvideo,esser2023structure, zhang2024show,zhang2023i2vgen,wang2024recipe,yuan2024instructvideo,wang2024videocomposer,bar2024lumiere,gupta2025photorealistic} have focused on improving traditional T2I models by incorporating temporal blocks and training these new components to convert T2I models into T2V models. Notable examples include AnimateDiff~\cite{guo2023animatediff}, Emu video~\cite{girdhar2023emu}, PYoCo~\cite{ge2023preserve}, and Align your Latents~\cite{blattmann2023align}. Furthermore, approaches such as LVDM~\cite{he2022latent}, VideoCrafter~\cite{chen2023videocrafter1,chen2024videocrafter2}, ModelScope~\cite{wang2023modelscope}, LAVIE~\cite{wang2023lavie}, and VideoFactory~\cite{wang2023videofactory} have utilized similar architectures, initializing with T2I models, and fine-tuning both spatial and temporal blocks to achieve enhanced visual outcomes. Besides, Sora~\cite{videoworldsimulators2024}, CogVideoX~\cite{yang2024cogvideox}, Latte~\cite{ma2024latte} and Allegro~\cite{zhou2024allegro} have made notable strides in video generation by integrating Transformer-based backbones~\cite{yu2024efficient, ma2024latte} and employing 3D-VAE technology. The development of these foundational models lays a solid foundation for customized video generation.

\subsection{Customized Image/Video Generation}
Similar to the development history of foundational models, the rapid advancement of text-to-image technology has spurred significant progress in customized generation within the image domain. Customized image generation, which adapts to user preferences, has attracted increasing attention~\cite{chen2023disenbooth, han2023svdiff, wei2023elite, shi2023instantbooth, li2024stylegan, ruiz2024hyperdreambooth, hua2023dreamtuner, han2024face, wang2024instantid, liu2023cones, liu2023cones2, chen2023anydoor,huang2024chatdittrainingfreebaselinetaskagnostic,liang2024ideabenchfargenerativemodels}. These works can be broadly categorized into two types based on whether the entire model needs to be retrained when changing the subject.
The first category includes methods such as Textual Inversion~\cite{gal2022image}, DreamBooth~\cite{ruiz2023dreambooth}, Custom Diffusion~\cite{kumari2023multi}, and Mix-of-Show~\cite{gu2024mix}. These approaches achieve full customization by learning a text token or directly fine-tuning all or part of the model's parameters. Although these methods often produce content with high visual fidelity to the specified subject, they require retraining when the subject changes.
The second category includes methods like IP-Adapter~\cite{ye2023ip}, InstantID~\cite{wang2024instantid}, and PhotoMaker~\cite{li2024photomaker}. These approaches employ various information injection techniques and leverage large-scale training to eliminate the need for parameter retraining when the subject changes.
Building on these methods, customized video generation has also evolved with advancements in foundational models.
DreamVideo~\cite{wei2023dreamvideo}, CustomVideo~\cite{wang2024customvideo}, Animate-A-Story~\cite{he2023animate}, Still-Moving~\cite{chefer2024still}, CustomCrafter~\cite{wu2024customcrafter}, and VideoAssembler~\cite{zhao2023videoassembler} achieve customization by fine-tuning parts of the Video Diffusion Model. However, this entails a higher training cost for users than customized image generation, resulting in significant inconvenience.
Some works, such as VideoBooth~\cite{jiang2024videobooth} and ID-Animator~\cite{he2024id}, attempt to adopt training methods similar to IP-Adapter. 
However, they have not yet achieved the same level of success as customized image generation.
% 一方面原因是受限于视频的基础模型和训练数据匮乏，一方面也有可能是信息注入的方式过于粗粒度，导致本就不强的text-to-video模型无法很好的学习到。
\section{Preliminary}
\textbf{Video diffusion models (VDMs)}~\cite{wang2023modelscope,he2022latent, guo2023animatediff, chen2024videocrafter2} are designed for video generation tasks by extending image diffusion models to adapt to video data. VDMs learn a video data distribution by the gradual denoising of a variable sampled from a Gaussian distribution. 
% This process simulates the reverse process of a fixed-length Markov Chain. 
First, a learnable autoencoder (consisting of an encoder ${\cal E}$ and a decoder ${\cal D}$) is trained to compress the video into a smaller latent space representation.
Then, a latent representation $z = {\cal E}(x)$ is trained instead of a video $x$.
Specifically, the diffusion model $\epsilon_{\theta}$ aims to predict the added noise $\epsilon$ at each timestep $t$ based on the text condition $c_{text}$, where $t \in \mathcal{U}(0, 1)$. The training objective can be simplified as a reconstruction loss:
\begin{equation}
    \mathcal{L}_{video} = \mathbb{E}_{z, c, \epsilon \sim \mathcal{N}(0, \mathbf{\mathrm{I}}), t}\left[\left\| \epsilon - \epsilon_\theta\left(z_t, c_{text} , t\right)\right\|_2^2\right],
\label{eq:diffusion_loss}
\end{equation}
where $z \in \mathbb{R}^{F \times H \times W \times C}$ is the latent code of video data with $F, H, W, C$ being frame, height, width, and channel, respectively.
$c_{text}$ is the text prompt for the input video.
A noise-corrupted latent code $z_t$ from the ground truth $z_0$ is formulated as $z_t = \lambda_t z_0 + \sigma_t \epsilon$, where $\sigma_t = \sqrt{1 - \lambda_t^2}$, $\lambda_t$ and $\sigma_t$ are hyperparameters to control the diffusion process. 
In this work, we selected the AnimateDiff~\cite{guo2023animatediff} as our base video diffusion model.
\begin{figure*}[tb]
    \centering
    \includegraphics[width=0.95\linewidth]{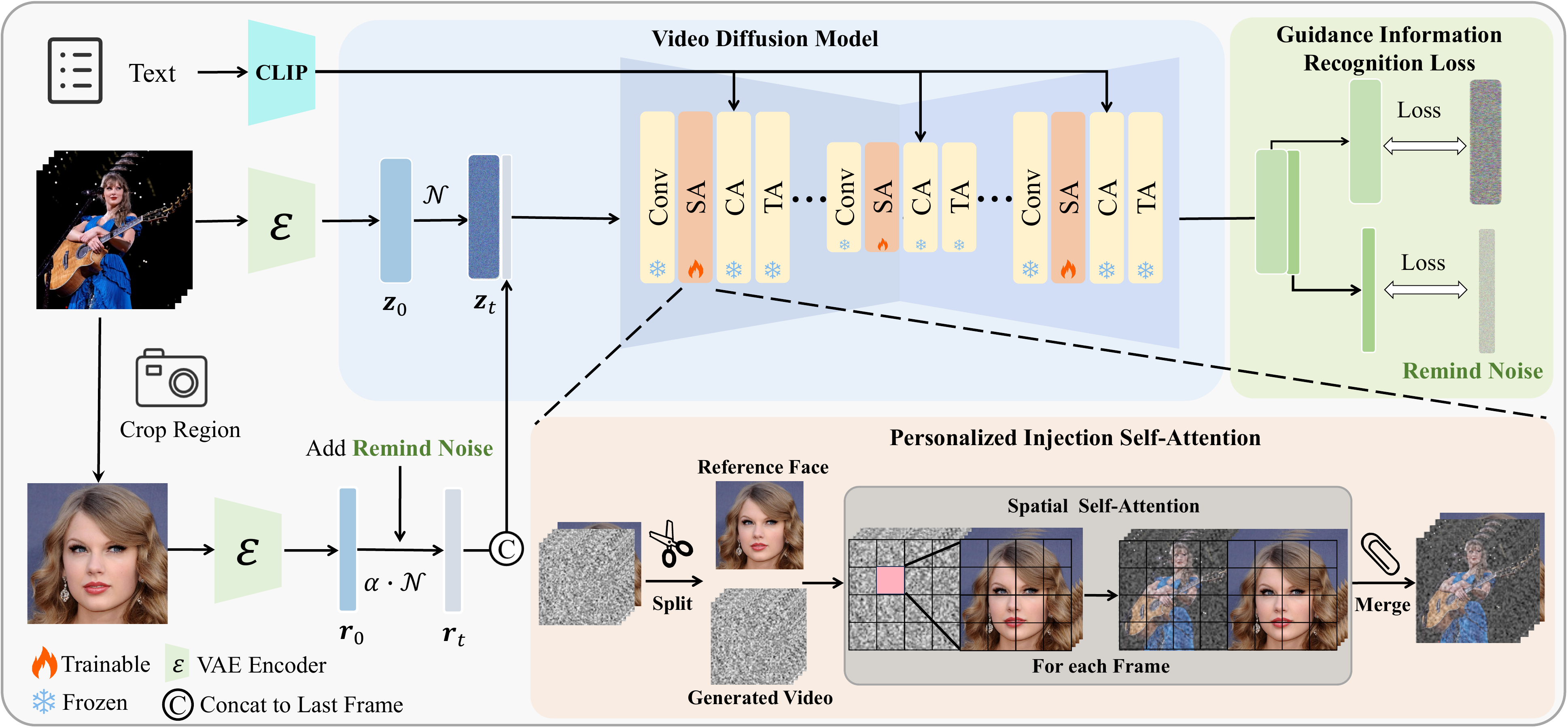}
    % For feature extraction, we directly input reference image into VDM, 利用VDM现有模块进行细粒度的特征提取。我们基于VDM中self-attention的功能，重新设定了其计算方式来完成feature injection。此外，为了更好的让模型辨识参考信息和生成内容，我们设计欧了Guidance Information Recognition Loss。
    \caption{Overall pipeline of VideoMaker. We directly input the reference image into VDM and use VDM's modules for fine-grained feature extraction. We modified the computation of spatial self-attention to enable feature injection. Additionally, to distinguish between reference features and generated content, we designed the Guidance Information Recognition Loss to optimize the training strategy.}
    \label{fig:pipline}
\end{figure*}
\section{Method}
% 给定一个subject的照片，我们的目标是让模型学习到这个人物的外貌以及物品属性。我们希望得到一个在更换指定subject后无需重新微调参数的模型，从而方便使用者方便的生成任何他们想要生成subject的视频。不同于简单的图生视频任务需要保持生成视频与参考图完全相似，我们希望模型能只学习到参考图像中主体的的概念和特征，并根据文本提示生成多样的视频。一个广泛的思路是通过提取subject特征，并向Video Diffusion Model注入所提取到的信息来完成这一任务。不同于先前的工作需要依赖于引入额外的图像编码器并依赖于简单的特征叠加或者是通过Cross-Attention注入参考特征，我们的方法不引入任何额外的模块，利用扩散模型自身作为特征提取器, 并利用模型内Self-Attention的作用机理来完成信息注入。我们在~\Cref{sebsec:explore}介绍了我们的方法的动机和核心思路，在\label{subsec:pipline}介绍了我们具体如何利用VDM来完成subject特征提取以及让VDM学习到subject的方法。此外，在~\Cref{subsec:loss}介绍了我们为了让模型更好的区分参考信息与生成内容区别所提出的训练策略。
Given a photo of a subject, our goal is to train a model that can extract the subject’s appearance and generate a video of the same subject.
Besides, the model does not require retraining when changing the subject.
% We utilize the VDM itself as a feature extractor and leverage the self-attention mechanism within the model to achieve information injection.
We discuss the key ideas of our method in ~\Cref{subsec:explore}, and detail how we utilize the inherent force of VDM to extract subject features and enable VDM to learn the subject in~\Cref{subsec:pipline}. 
In~\Cref{subsec:loss}, we introduce our proposed training strategy to better distinguish between reference information and generated content.
Furthermore, we add some details about the training and inference in \Cref{subsec:paradigm}.

\subsection{Explore Video Diffusion Model}
\label{subsec:explore}

To achieve customized video generation, two core problems must be addressed: subject feature extraction and feature injection.
For subject feature extraction, some works employ cross-model alignment models, such as CLIP~\cite{radford2021learning}. 
However, because of their training tasks, these models produce coarse-grained features that fail to capture the subject's appearance in detail.
Some studies attempt to train a ReferenceNet but significantly increase training overhead. 
We propose a new method leveraging the pre-trained VDM for subject feature extraction.
When the subject reference image is input directly into the VDM without added noise, this can be considered as a special case of VDM at $t=0$.
Therefore, the VDM can accurately process and extract the features of the noise-free reference image.
This approach allows for extracting fine-grained subject features without additional training overhead while reducing the domain gap between the extracted features and the VDM's inherent knowledge.
% This approach enables the extraction of fine-grained subject features without additional training costs and minimizes the domain gap between the extracted features and the VDM’s inherent knowledge.

Regarding feature injection, spatial cross-attention is used for VDM's cross-modal interaction between image and text. 
Influenced by this design, existing methods employ cross-attention heuristically to inject subject features. 
However, spatial self-attention in VDM is responsible for modeling the relationships between pixels within a frame. 
In customized video generation, a key objective is to ensure the subject ``appears" in the frame.
So, injecting subject features when constructing pixel relationships within the frame is a more direct method. 
Moreover, spatial self-attention can selectively interact with these features, which helps promote diversity in the generated videos.
Benefiting from the feature extraction performed by the VDM itself, we can directly use the VDM's inherent spatial self-attention modeling capability for more direct information interaction.

\subsection{Personalized Injection Self Attention}
\label{subsec:pipline}
% 为了让模型学习到subject的外貌信息和属性，我们要首先要对subject的特征进行提取。与前人不同的是，我们利用Diffusion自身已有的网络结构来完成这个过程。
\noindent\textbf{Subject feature extraction.} 
% To enable the model to learn the appearance and attributes of the subject, we first extract its features. 
Unlike previous approaches, we leverage the existing network structure of the VDM to achieve this, \ie Resblock in unit-based VDM.
% 如\Cref{fig:pipline}所示，给定一个视频$x$通过VAE得到的latent space编码并加噪的结果$Z_{t} \in \mathbb{R}^{F \times H \times W \times C}$,以及一张指定subject的参考图R，我们首先将参考图$R$通过VAE编码得到$r$但不加噪。然后，我们将得到的参考图latent space的编码$r$与$Z_{t}$在frame的维度上进行拼接，从而得到$z_t^{prime} \in \mathbb{R}^{F+1 \times H \times W \times C$作为模型的实际输入。
As illustrated in ~\Cref{fig:pipline}, given a video $x$ that is encoded into the latent space and then noised to obtain $z_t \in \mathbb{R}^{F \times H \times W \times C}$ through a VAE, along with a reference image $R$ of the specified subject, we first encode the reference image $R$ using the VAE to obtain $r$ without adding noise. 
We then concatenate the encoded reference image latent space $r$ with $z_t$ along the frame dimension, resulting in $ z_t^{\prime} \in \mathbb{R}^{(F+1) \times H \times W \times C}$ as the actual input to the model.
% 然后，我们利用VDM自身的ResBlock作为特征提取器来对$z_t^{prime}$进行特征提取，得到Self-Attention层的输入$f\in \mathbb{R}^{F+1 \times h \times w \times c$。然后我们将特征$f$进行分离，分别得到待生成视频对应的噪声部分$f_z \in \mathbb{R}^{F \times h \times w \times c$和参考信息所对应的$f_r \in \mathbb{R}^{F \times h \times w \times c$部分.到此，我们完成了对指定subject的特征提取。
Next, we use the Resblock as a feature extractor to extract features from $z_t^{\prime}$, obtaining the input $f \in \mathbb{R}^{(F+1) \times h \times w \times c}$ for the spatial self-attention layer. 
We then separate the features $f$ to obtain the noise part corresponding to the video to be generated $f_z \in \mathbb{R}^{F \times h \times w \times c}$ and the part corresponding to the reference information $f_r \in \mathbb{R}^{1 \times h \times w \times c}$.
We have completed the feature extraction for the specified subject at this stage.

% 在得到指定subject特征后，我们接下来要将这个特征注入到VDM中去。对于$f_z$中的任一一帧$f_z^i$，在计算Self-Attention前会将其转变为$h \times w$个token。我们将$f_r$与$f_z^i$进行拼接，从而使得Self-Attention层每一帧的输入变为$2 \times h \times w$个token, 我们用$\mathbf{X}$来代表这些token.然后，我们通过Self-Attention将信息进行融合：
% 其中$\mathbf{X}^{\prime}$为输出的attention特征.$\mathbf{Q}$,$\mathbf{K}$,$\mathbf{V}$分别代表query,key,value数组。具体而言，$\mathbf{Q}=\mathbf{X} \mathbf{W}_Q, \mathbf{K}=\mathbf{X} \mathbf{W}_K, \mathbf{V}=\mathbf{X} \mathbf{W}_V$.$\mathbf{W}_Q$,$\mathbf{W}_K$,$\mathbf{W}_V$是对应的projection matrices, d is the dimension of key features.

\noindent\textbf{Subject feature injection.} After extracting the specified subject features, injecting these features into the VDM is next. For each frame $f_z^i$ in $f_z$, it is transformed into $h \times w $ tokens before computing spatial self-attention. We concatenate $f_r $ with $f_z^i $ so that the input to the spatial self-attention layer for each frame becomes $2 \times h \times w $ tokens. We denote these tokens as $\mathbf{X} $. Then, we fuse the information through spatial self-atention:
\begin{equation}
    \mathbf{X}^{\prime} = \operatorname{Attention}(\mathbf{Q}, \mathbf{K}, \mathbf{V}) = \operatorname{Softmax}\left(\frac{\mathbf{Q} \mathbf{K}^{\top}}{\sqrt{d}}\right) \mathbf{V}
\end{equation}
where $\mathbf{X}^{\prime} $ represents the output attention features, $\mathbf{Q} $, $\mathbf{K} $, and $\mathbf{V} $ represent the query, key, and value matrices, respectively. Specifically, $\mathbf{Q} = \mathbf{X} \mathbf{W}_Q $, $\mathbf{K} = \mathbf{X} \mathbf{W}_K $, and $\mathbf{V} = \mathbf{X} \mathbf{W}_V $. $\mathbf{W}_Q $, $\mathbf{W}_K $, and $\mathbf{W}_V $ are the corresponding projection matrices, and $d$ is the dimension of the key features.
After computing the attention, we separate the output attention features $\mathbf{X}^{\prime}$ to obtain $f_z^{\prime}$ and $f_r^{\prime}$. Since $f_r^{\prime}$ is repeated $F$ times, we take the average of the $F$ corresponding results as the final $f_r^{\prime}$. Finally, we concatenate the obtained $f_r^{\prime}$ with $f_z^{\prime}$ to obtain the updated $f^{\prime}$, which is then fed into the subsequent model layers for further processing.

\subsection{Guidance Information Recognition Loss}
\label{subsec:loss}
Since the actual input $z_t^{\prime}$ to our framework includes an additional frame compared to the input in Equation~\ref{eq:diffusion_loss}, the output $\epsilon_{\theta} \in \mathbb{R}^{(F+1) \times H \times W \times C}$ also has an extra frame relative to the output in Equation~\ref{eq:diffusion_loss}. 
A straightforward training objective would be to eliminate the output corresponding to the reference information $r$ and compute the loss only for the remaining frames. 
This approach encourages the model to focus on learning customized video generation with the specified subject.
% This approach allows the model to focus on learning the customized generation of videos with the specified subject.
However, our observations of the final results revealed that without supervision on the reference information, the model struggles to accurately recognize that the reference information $r$ is an image without adding noise, leading to instability in the generated results.
To address this, we introduced a Guidance Information Recognition Loss to supervise the reference information, enabling the model to accurately distinguish between the reference information and the generated content, thereby improving the quality of the customized generation.
Specifically, during the training process at timestep $t$, we add a remind noise to the reference information $r$:
\begin{equation}
    r_{t} = \lambda_{{t^{\prime}}} r + \sqrt{1 - \lambda_{{t^{\prime}}}^2} \epsilon ,
    \label{eq:ref_noise}
\end{equation}
where $t^{\prime} = \alpha \cdot t$, and $\alpha$ is a manually set hyperparameter. 
% 注意，为了防止额外加入的噪声对参考信息造成较大破坏，$\alpha$值设定一个较小值，来确保remind noise是一个轻微的噪声。
To prevent the added noise from heavily degrading the reference information, $\alpha$ is set to a small value to ensure that the reminder noise remains minimal.
When computing the loss function, we also calculate the loss same as ~\Cref{eq:diffusion_loss} for the reference information $r$:
\begin{equation}
    \mathcal{L}_{reg} = \mathbb{E}_{r, c, \epsilon \sim \mathcal{N}(0, \mathbf{\mathrm{I}}), t}\left[\left\| \epsilon - \epsilon_\theta\left(r_t, c_{text}, t\right)\right\|_2^2\right].
\label{eq:reg_loss}
\end{equation}
We use $\mathcal{L}_{reg}$ as an auxiliary optimization objective, combined with the main objective, to guide the model’s training:
\begin{equation}
    \mathcal{L} = \mathcal{L}_{video} + \beta \cdot \mathcal{L}_{reg} ,
\label{eq:all_loss}
\end{equation}
where $\beta$ is a hyperparameter. 
To avoid interfering with the optimization of the main customized video generation task, $\beta$ is chosen as a relatively small value.

\subsection{Training and Inference Paradigm}
\label{subsec:paradigm}
\noindent\textbf{Training.} 
Our framework’s straightforward design allows us to avoid the need for additional subject feature extractors during training. 
Since we only adjust the number of input tokens to the model’s original spatial self-attention layer, injecting subject information into the VDM does not increase the parameter count. 
We assume that the ResBlock in pretrained VDM is already sufficient to extract the feature information from the reference image. 
Therefore, our model needs to fine-tune the original VDM's spatial self-attention layer while freezing the parameters of the remaining parts during training. 
In addition, To enable temporal attention todistinguish well between reference information and generated videos, we recommend fine-tuning the parameters of the motion block synchronously during training.
It can also achieve customized video generation without fine-tuning the motion blocks.
We also randomly drop image conditions in the training stage to enable classifier-free guidance in the inference stage:
\begin{equation}
    \hat{\boldsymbol{\epsilon}}_\theta\left(\boldsymbol{z}_t, \boldsymbol{c}_t, r, t\right)=w \boldsymbol{\epsilon}_\theta\left(\boldsymbol{z}_t, \boldsymbol{c}_t, r, t\right)+(1-w) \boldsymbol{\epsilon}_\theta\left(\boldsymbol{z}_t, t\right).
\end{equation}

\noindent\textbf{Inference.}
During the inference process, for the output of the model, the output corresponding to the reference information is discarded directly.
Additionally, although we added light noise to the subject's reference image during training to explicitly help the model distinguish the guidance information, we chose to remove the noise addition to the reference image during inference. This ensures that the generated video is not affected by noise, thus maintaining the quality and stability of the output.
\section{Experiments}
\subsection{Experimental Setup}
\label{sc:exp_set}
\noindent\textbf{Datasets.}
% 由于现有大规模通用物品视频定制化数据集的缺乏，我们选用了CelebV-Text~\cite{yu2023celebv}和VideoBooth~\cite{jiang2024videobooth}两个数据集分别进行定制化Human Video生成、和定制化物品生成实验。 CelebV-Text数据集拥有70000条分辨率至少为$512 \times 512$人脸面部视频片段以及半自动标注的文本描述，视频总时长约279hrs.为了使得模型专注于ID的学习，避免过拟合subject参考图像的背景，我们首先使用Grounding DINO~\cite{liu2023grounding}以‘person’作为提示词对是每个视频随机采样的4帧进行处理，得到每个视频中人所对应边界框，再融入SAM模型得到subject mask，将mask以外区域置为白色，作为我们每个视频的参考图像。在训练时候，我们随机选用4帧中其中任意一帧作为实际输入的参考图像。
% Videobooth Datasets是由从WebVid~\cite{bain2021frozen}的子集中筛选出48724条视频数据组成，包含cat,dog,car等九类物品，由于其数据中提供了每个视频所对应subject的参考图像，我们直接使用了该数据集提供的参考图作为实际输入的参考图。
Due to the lack of large-scale video customization datasets, we selected CelebV-Text~\cite{yu2023celebv} and VideoBooth datasets~\cite{jiang2024videobooth} for our experiments on customized human video generation and customized object generation, respectively.
The CelebV-Text dataset contains 70,000 facial video clips with a resolution of at least $512 \times 512$ and semi-automatically annotated text prompts. 
To ensure that the model focuses on learning the subject and avoids overfitting to the background of the subject's reference image, we use subject highlight preprocessing same as VideoBooth. 
We use Grounding DINO~\cite{liu2023grounding} and~\cite{kirillov2023segment} to remove the background and get the main subject of a frame, which we then use as the reference image.
The VideoBooth dataset consists of 48,724 video clips selected from a subset of WebVid~\cite{bain2021frozen}, covering nine categories of objects: bear, car, cat, dog, elephant, horse, lion, panda, tiger. 
Since each video in this dataset is accompanied by a reference image, we use the provided reference images directly as input.

\noindent\textbf{Implementation details.}
% 为了方便与其他现有方法做比较，我们fellow了ID-Animator工作的设定，使用AnimateDiff的Stable Diffusion 1.5版本作为基础模型进行实验。所有的实验均使用4张NVIDIA A100进行，每张卡的batchsize设定为1，我们固定每个视频的frame stride为8，并设定输入视频大小为$512 \times 512$。我们使用AdamW优化器，将学习率设定为$1 \times 10^{-5}$, weight decay设定为$1 \times 10^{-2}$。对于在CelebV-Text数据集上进行的定制化Human Video生成实验，我们训练150000步，对于在Videobooth数据集上进行的实验，由于数据规模较小，我们修改迭代步数为100000步。
To facilitate comparison with other existing methods, we followed the setup of~\cite{he2024id,ye2023ip} and used the Stable Diffusion 1.5 version of AnimateDiff as the base model for our experiments. 
All experiments are conducted using four NVIDIA A100 GPUs, with a batch size of 1 per GPU. 
We fixed the frame stride for each video at 8 and set the output video resolution to $512 \times 512$.
We used the AdamW optimizer, setting the learning rate to $1 \times 10^{-5}$ and the weight decay to $1 \times 10^{-2}$. 
For the customized human video generation experiments on the CelebV-Text dataset, we trained for 150,000 steps. 
For the VideoBooth dataset, due to the smaller data scale, we adjusted the training to 100,000 steps.
For the hyperparameter settings in our method, we set $\alpha$ in \Cref{eq:ref_noise} to 0.01, and $\beta$ in \Cref{eq:all_loss} to 0.1.
During the inference generation process, we use DDIM~\cite{song2020denoising} for 30-step sampling and a classifier-free guidance scale of 8 to generate videos.

\noindent\textbf{Baselines.}
Since customized human video generation is significantly more challenging than customized object generation and better highlights the capability of customized generation, we primarily focus on comparing customized human video generation to demonstrate the effectiveness of our method.
We selected IP-Adapter~\cite{ye2023ip}, IP-Adapter-Plus, IP-Adapter-FaceID and ID-Animator~\cite{he2024id} for a fair comparison. 
IP-Adapter-Plus represents an enhanced version of IP-Adapter that utilizes Q-Former~\cite{li2023blip} to extract features from CLIP image embeddings, while IP-Adapter-FaceID substitutes CLIP with a dedicated face recognition model.
Since PhotoMaker~\cite{li2024photomaker} only has pretrained weights for the SDXL~\cite{podell2023sdxl} version, we used the results generated with AnimateDiff SDXL at a resolution of $512 \times 512$ for comparison. 
For customized object video generation, we use VideoBooth~\cite{jiang2024videobooth} as the baseline for comparison.

\noindent\textbf{Evaluation metrics.}
% Follow~\cite{wei2023dreamvideo,wang2024customvideo,he2024id}, 我们从overall consistency, subject fidelity两个方面对方法生成视频质量进行评估。
% 对于overall consistency方面，我们选用CLIP image-text similarity(CLIP-T)、Temporal Consistency、Dynamic Degree(DD) 三个指标来进行比较.
% CLIP-T calculates the average cosine similarity between CLIP~\cite{radford2021learning} image embeddings of all generated frames and their text embedding.
% Temporal Consistency compute CLIP image embeddings on all generated frames and report the average cosine similarity between all pairs of consecutive frames.
% DD uses optical flow to measure motion dynamics.
% 对于subject fidelity方面测评，对于定制化human video生成以及定制化物品生成，我们都使用了CLIP-I和DINO-I进行了测评：
% CLIP-I measures the visual similarity between the generated and target subjects. We computed the average cosine similarity between the CLIP image embeddings of all generated frames and the target images.
% DINO-I~\cite{ruiz2023dreambooth}, another metric for measuring the visual similarity using ViTS/16 DINO~\cite{zhang2022dino}. Compared to CLIP, the self-supervised training model encourages the distinguishing features of individual subjects.
% 此外，对于定制化human video生成，由于CLIP-I所提取特征较粗，我们还引入了Face Similarity~\cite{deng2019arcface}来进行细粒度的精确比较，更精准的比对我们方法和baseline的性能。
Following the ~\cite{wei2023dreamvideo, wang2024customvideo, he2024id}, we evaluate generated video quality from two perspectives: overall consistency and subject fidelity.
We employ three metrics for overall consistency: CLIP-T, Temporal Consistency (T. Cons.), and Dynamic Degree (DD).
CLIP-T measures the average cosine similarity between the CLIP~\cite{radford2021learning} image embeddings of all generated frames and their text embeddings.
T. Cons. calculates the average cosine similarity between the CLIP image embeddings of consecutive frames.
DD~\cite{huang2024vbench} utilizes optical flow to quantify motion dynamics.
To evaluate subject fidelity, we use CLIP-I and DINO-I for both customized human and object video generation tasks. 
CLIP-I assesses the visual similarity between the generated frames and the target subjects by computing the average cosine similarity between the CLIP image embeddings of all generated frames and the reference images.
DINO-I~\cite{ruiz2023dreambooth} is another metric for visual similarity, using ViT-S/16 DINO~\cite{zhang2022dino}. 
Additionally, for customized human video generation, since CLIP-I captures relatively coarse visual features, we also incorporate Face Similarity~\cite{deng2019arcface} for a more fine-grained and precise comparison, enhancing the accuracy of our subject fidelity assessment.
\subsection{Quantitative Comparison}
\label{sc:cmp_value}
\begin{table}
    \centering
    \resizebox{\linewidth}{!}{
    \begin{tabular}{lcccccc}
    \toprule
    Method            & CLIP-T          & Face Sim.       & CLIP-I          & DINO-I          & T.Cons.         & DD              \\
    \midrule
    IP-Adapter        & 0.2064          & 0.1994          & 0.7772          & 0.6825          & \underline{0.9980}    & 0.1025          \\
    IP-Adapter-Plus   & 0.2109          & 0.2204          & \underline{0.7784}    & \underline{0.6856}    & \textbf{0.9981}          & 0.1000          \\
    IP-Adapter-Faceid & 0.2477          & \underline{0.5610}    & 0.5852          & 0.4410          & 0.9945          & 0.1200          \\
    ID-Animator       & 0.2236          & 0.3224          & 0.4719          & 0.3872          & 0.9891 & 0.2825          \\
    Photomaker(SDXL)  & \textbf{0.2627} & 0.3545          & 0.7323          & 0.4579          & 0.9777          & \underline{0.3675}    \\
    Ours              & \underline{0.2586}    & \textbf{0.8047} & \textbf{0.8285} & \textbf{0.7119} & 0.9818          & \textbf{0.3725} \\
    \bottomrule
    \end{tabular}
    }
    \caption{Comparison with the existing methods for customized human video generation. The best and the second-best results are denoted in bold and underlined, respectively.}
    \label{tab:sota}
\end{table}
\begin{table}[]
    \centering
    \resizebox{\linewidth}{!}{
    \begin{tabular}{lccccc}
    \toprule
    Method     & CLIP-T         & CLIP-I          & DINO-I          & T.Cons.         & DD              \\
    \midrule
    VideoBooth & 0.266          & 0.7637          & 0.6658          & 0.9564          & 0.5091          \\
    Ous        & \textbf{0.284} & \textbf{0.8071} & \textbf{0.7326} & \textbf{0.9848} & \textbf{0.5132} \\
    \bottomrule
    \end{tabular}
    }
    \caption{Comparison with the existing methods for customized object video generation}
    \label{tab:sota2}
\end{table}

\begin{figure}[tb]
    \centering
    \includegraphics[width=1.0\linewidth]{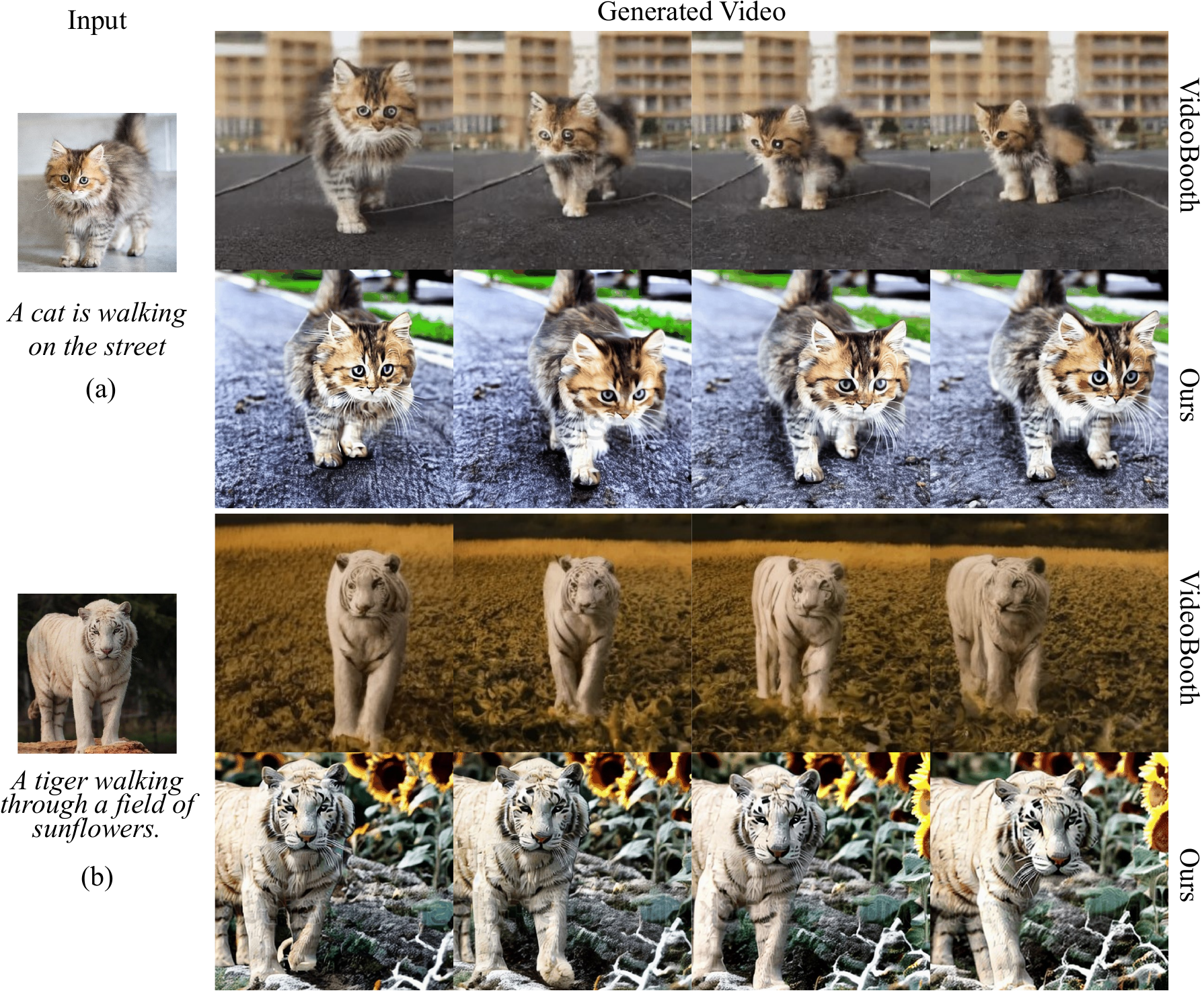}
    \caption{Qualitative comparison for customized object video generation. Compared with the blurry videos generated by VideoBooth~\cite{jiang2024videobooth}, our generated videos have more details.}
    \label{fig:cmp_object}
\end{figure}
\noindent\textbf{Customized object video generation.}
\begin{figure*}[tb]
    \centering
    \includegraphics[width=1.0\linewidth]{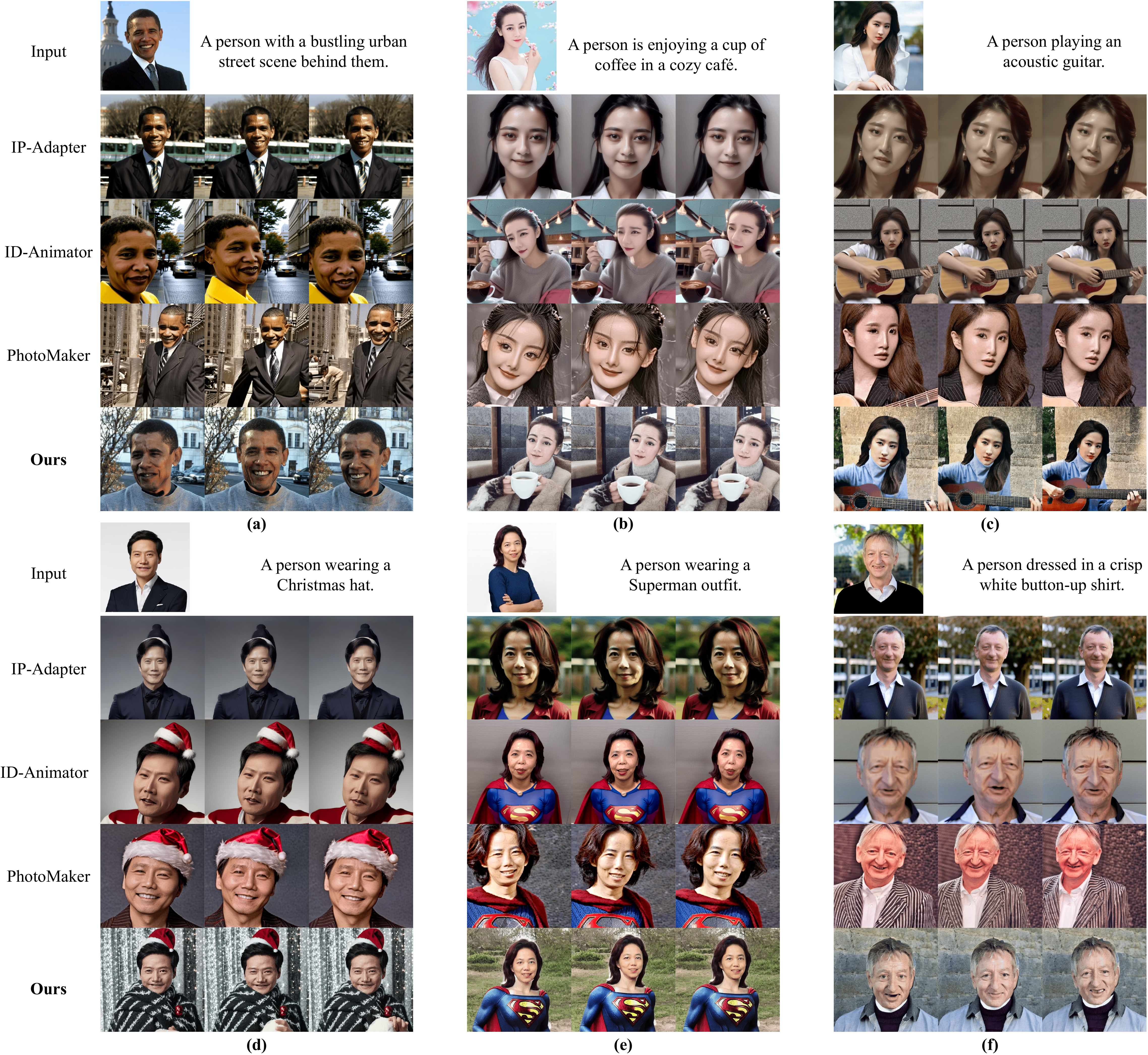}
    % 我们利用VDM自身的
    \caption{Qualitative comparison for customized human video generation. We compare our method with IP-Adapter~\cite{ye2023ip}, ID-Animator~\cite{he2024id} and PhotoMaker~\cite{li2024photomaker}. We observe that our method achieves high-quality generation, promising editability, and subject fidelity.}
    \label{fig:cmp_human}
\end{figure*}
\begin{table*}
    \centering
    \begin{tabular}{ccccccccccc}
    \toprule
    PISA                      & GIRL                      & W/O Cross                 & Update Motion                & SHP                      & CLIP-T & Face Sim. & CLIP-I & DINO-I & T.Cons. & DD     \\
    \midrule
    \checkmark &                           &                           &                           &                           & 0.2206 & 0.7928    & 0.7966 & 0.6694 & 0.9671  & 0.2725 \\
    \checkmark & \checkmark &                           &                           &                           & 0.2258 & 0.8184    & 0.8484 & 0.7536 & 0.9855  & 0.2750  \\
    \checkmark & \checkmark & \checkmark &                           &                           & 0.2291 & 0.8454    & 0.8469 & 0.7351 & 0.9747  & 0.2915 \\
    \checkmark & \checkmark & \checkmark & \checkmark &                           & 0.2302 & 0.8563    & 0.8674 & 0.7635 & 0.9823  & 0.3575 \\
    \checkmark & \checkmark & \checkmark & \checkmark & \checkmark & 0.2586 & 0.8047    & 0.8285 & 0.7119 & 0.9818  & 0.3725 \\
    \bottomrule
    \end{tabular}
    \caption{Quantitative results of each component. ``PISA" is our Personalized Injection Self Attention, GIRL is our Guidance Information Recognition Loss, ``W/O Cross" refers to whether our reference frame interacts with the text prompt, ``Update Motion" refers to whether to update the motion block, ``SHP" is our subject highlight preprocessing for datasets, }
    \label{tab:abalation}
\end{table*}
% 对于定制化物品视频生成，我们为VideoBooth数据集所包含的9个类别的物品，每个类别选取2个数据集外的样例,共计18个subject。对于每个subject, 我们根据其类别用ChatGPT生成10个prompt进行测试。
% 如~\Cref{tab:sota2}所示，相比于VideoBooth这类方法，我们的方法在text alignment,subject 外观一致性，生成视频跨帧一致性以及动态性都取得了更好的效果
We selected two examples for each of the nine object categories included in the VideoBooth dataset, totaling 18 subjects. For each subject, we generated 10 prompts tailored to their category using ChatGPT~\cite{achiam2023gpt}.
As shown in~\Cref{tab:sota2}, our method outperforms VideoBooth across all evaluation metrics, demonstrating the effectiveness of our approach.

\noindent\textbf{Customized human video generation.} 
% Follow~\cite{he2024id,li2024photomaker,wang2024instantid}, 我们收集了20个不同的人作为测试benchmark。对于每个人，我们使用ChatGPT~\cite{achiam2023gpt}生成了25个prompt，涵盖expressions, attributes, decorations, actions, and backgrounds五个方面来进行全面的测评。 
% \Cref{tab:sota}展示了我们测评的结果。
% 表中的IP-Adapter-Plus是IP-Adapter使用Q-Former to extract face features from CLIP image embeddings的版本，IP-Adapter-FaceID是使用人脸判别模型代替CLIP.
% 可以看到，我们的方法在Face Similarity、CLIP-I、DINO-I都明显优于现有方法。特别是Face Similarity这种细粒度的衡量生成内容与subject的外观一致性的指标，我们的方法展现出极强的性能。这证明了我们的这种利用模型自身能力进行定制化生成的方法能够更好的提取subject的外貌特征，将特征信息注入到VDM中。
% 此外，在text alignment方面，我们的方法在同基础模型上取得了最优的结果。PhotoMaker的基础模型是AnimateDiff SDXL version，在基础模型层面拥有更强大的生成能力，但我们的方法能够取得较为接近的成果，这说明了我们采用模型原生能力进行subject信息注入的方式能在实现高保真的subject外观一致性的基础上，保证生成视频与给定prompt的匹配。
% 我们的方法落后于IP-Adapter系列模型的主要原因是这些模型生成的视频趋近于静止，因此拥有较高的跨帧一致性。Dynamic Degree指标一栏说明了这个问题。此外，我们的方法在生成内容的动态上也取得了较好的效果。
Following~\cite{he2024id,li2024photomaker,wang2024instantid}, we created a testing benchmark comprising 16 different individuals.
% We generated 25 prompts for each person using ChatGPT, covering five aspects: expressions, attributes, decorations, actions, and backgrounds for comprehensive evaluation.
we generated 25 prompts using ChatGPT, addressing five aspects: expressions, attributes, decorations, actions, and backgrounds, to enable a comprehensive evaluation. 
All images and prompts are provided in the Appendix.
As shown in \Cref{tab:sota}, our method significantly outperforms existing methods in Face Similarity, CLIP-I, and DINO-I. 
Our method demonstrates strong performance, especially for Face Similarity, a fine-grained metric that measures subject fidelity. 
This proves that our approach of using the model's inherent force for customized generation can better extract the subject's features and inject these features into the VDM.
For text alignment, our method achieved the best results on the same base model. 
% The base model of PhotoMaker is the AnimateDiff SDXL, which has more powerful generation abilities at the base model level. 
% However, our method can achieve relatively close results, which shows that our methods can achieve better text alignment while maintaining high subject fidelity.
While PhotoMaker’s base model, AnimateDiff SDXL, possesses stronger generative capabilities at the base model level, our method achieves comparable results, demonstrating improved text alignment while maintaining high subject fidelity.
The main reason for our T. Cons. behind the IP-Adapter series models is that the videos generated by these models tend to remain static, leading to higher cross-frame consistency. The DD metric illustrates this issue. 
Moreover, our method has a better degree of dynamicity.

\subsection{Qualitative Comparison}
% 为了进一步验证我们方法的有效性，我们可视化了我们的方法与现有的方法生成结果进行对比。
% for human generation, 如图~\Cref{fig:cmp_human}所示，我们的方法们可以在保证text alignment的同时，相比于其他的方法有着明显的提升subject fidelity。我们注意到，相比与其他方法，我们方法所生成的人往往具有更多的面部细节，凸显了我们利用VDM自身进行特征提取以及信息注入在保持subject外观一致性上的优势。
% for object generation, 如图~\Cref{fig:cmp_object}，相比于VideoBooth在生成结果中存在猫和老虎的面部细节缺失的问题，我们方法所生成的视频可以高保真的提取subject的纹理细节并生成。同时，VideoBooth在~\Cref{fig:cmp_object}(b)未能正确的根据prompt生成向日葵，而我们的方法可以，证明我们方法有更强的Text Alignment能力。注意，我们的生成的方法有水印是因为数据集中含有水印，Videobooth是通过在模型外额外外接了一个模型来去除水印。
To further validate the effectiveness of our method, we compared the visual results generated by our method with existing methods.
For the human generation, as shown in \Cref{fig:cmp_human}, our method significantly improves subject fidelity compared to other methods while ensuring text alignment.
Notably, our generated humans exhibit more refined facial details, underscoring the advantages of using VDM for feature extraction and information injection, contributing to enhanced consistency in subject appearance.
For object generation, as illustrated in ~\Cref{fig:cmp_object}, our method produces videos that faithfully capture the subject's texture details, whereas VideoBooth often loses facial details when generating animals like cats and tigers. Additionally, in ~\Cref{fig:cmp_object}(b), VideoBooth failed to generate a sunflower according to the prompt, whereas our method successfully rendered it.
Note that the watermark on our generated results comes from the dataset itself, while VideoBooth removes it using an additional external model.
% 我们提供了更多的可视化对比结果and demo video，可以在补充材料找到。
We provide more qualitative comparison results and their videos, which can be found in the supplementary materials.

\subsection{Ablation Studies}
% We designed detailed ablations on the CelebV-Text dataset to discuss the effectiveness of each component and verify some framework design details.
We conducted a series of ablation experiments on the CelebV-Text dataset to evaluate the effectiveness of each component in our framework and validate specific design choices. 
% In our framework, simply using the reference frame as the as input will continue to participate in the calculation of cross-attention and temporal-attention after participating in the self-attention. 
% We designed experiments to analyze the impact of participating in these blocks.
% In our model, the reference frame initially participates in self-attention before contributing to cross-attention and temporal-attention computations. 
In our model, the reference frame participates in cross-attention and temporal attention computations after participating in self-attention.
Our ablations examine the influence of each processing step on model performance.
Since part of the methods we are comparing (\eg IP-Adapter) do not perform additional data processing, for fairness, we conducted experiments using the original data in the first four rows of \Cref{tab:abalation}, using random frames of the video as the reference images.

\noindent\textbf{Effect of Personalized Injection Self Attention.}
In the initial experiment, we applied feature injection exclusively through self-attention while preventing the reference frame from influencing cross-attention calculations. This setup supervised only the frames corresponding to the generated video. 
As shown in the first line of \Cref{tab:abalation}, compared to existing methods, just by modifying the subject extraction and injection, we can significantly improve the subject fidelity.
% As shown in the first row of \Cref{tab:abalation}, our model significantly outperforms existing methods in maintaining subject fidelity, simply through enhanced subject extraction.

\noindent\textbf{Effect of Guidance Information Recognition Loss.}
To improve appearance consistency, we introduce Guidance Information Recognition Loss, designed to help the model accurately distinguish reference frame from other frames.
This component further stabilizes the subject's appearance across frames, as demonstrated in the second row of \Cref{tab:abalation}.

\noindent\textbf{Whether to participate in cross-attention.}
Allowing the reference frame to participate in cross-attention computation introduces two potential outcomes: (1) Altering the subject features, which could negatively impact subject fidelity, or (2) Enabling interaction with textual information, thereby improving text alignment in the generated video and enhancing the coherence of the reference features. 
To assess this, we conducted experiments where the reference frame was included in cross-attention. The results in row three indicate that allowing this interaction enhances text alignment without compromising subject fidelity.

\noindent\textbf{Whether Update Motion Blocks.}
As we mentioned in \Cref{subsec:paradigm}, we found that incorporating motion block training during fine-tuning helps the model better distinguish between reference information and generated video. 
This approach also enhances video dynamics without compromising subject fidelity, as shown in row four of \Cref{tab:abalation}.

\noindent\textbf{Effect of Subject Highlight Preprocessing.}
Finally, we evaluated the impact of subject-specific data preprocessing. Compared to using a random video frame as the subject reference, our preprocessing approach reduces the model’s tendency to overfit to irrelevant background details, directing focus onto the subject’s appearance features. 
This improved alignment between the generated video and text prompts, as seen in the last row of  \Cref{tab:abalation}.
\section{Conclusion}
\vspace{+0.1cm}
% 在本文中，我们提出了VideoMaker,一个全新的利用the Inherent Force of VDM来完成高质量zero-shot定制化生成框架。相比于启发式的外接模型来进行subject特征提取和注入，我们巧妙的使用VDM自身的力量完成了定制化生成所需的subject细粒度特征提取与注入。我们的模型在不添加任何额外的模块的情况下，实现了高质量的定制化视频生成。
In this paper, we propose VideoMaker, a novel framework that uses the inherent force of VDM to achieve high-quality zero-shot customized generation. 
Compared with the heuristic external model to extract and inject subject features, we discover and use the force of inherent VDM to complete the fine-grained subject feature extraction and injection required for customized generation.
% Our model achieves high-quality customized video generation without adding any additional modules.
Experimental results confirm the efficacy of our approach across both customized human and object video generation tasks.
{
    \small
    \bibliographystyle{ieeenat_fullname}
    \bibliography{main}
}

\clearpage
\setcounter{page}{1}
\setcounter{section}{0}
\setcounter{figure}{0}
\setcounter{table}{0}
\maketitlesupplementary
\appendix

\section{Dataset Details}
\paragraph{Training dataset.}
% 正如我们在正文Section 5.1所提到的，我们采用了subjecthighlightpreprocessing对数据集进行了处理。
% 具体而言，We first use Grounding DINO~\cite{liu2023grounding} with the prompt `person.' to process a randomly sampled frame from each video. 
% This provides the bounding box corresponding to the person in each video.
% We then integrate the SAM~\cite{kirillov2023segment} model to obtain the subject mask and set the area outside the mask to white, which serves as the reference image for each video. 
% During training, we randomly select any one of the four frames as the actual input reference image.
% 同时，我们在训练冲去除了视频中还有多个人或者人脸占比过小的视频数据，最后处理后的CelebV-Text数据集含有40600个视频。
% 此外，我们在训练时候采用了\textit{RandomHorizontalFlip}以及\textit{RandomAffine}作为数据增强。
As mentioned in \Cref{sc:exp_set} of the main text, we employed subject highlight preprocessing to process the dataset. Specifically, we first use Grounding DINO~\cite{liu2023grounding} with the prompt ``head" to process a randomly sampled frame from each video. This provides the bounding box corresponding to the person in each video. We then integrate the SAM~\cite{kirillov2023segment} model to obtain the subject mask and set the area outside the mask to white, which serves as the reference image for each video. During training, we randomly select any one of the four frames as the actual input reference image. 
Additionally, we removed videos containing multiple people or those where the proportion of the face is too small. After processing, the CelebV-Text dataset contains 40,600 videos. 
% Furthermore, we used \textit{RandomHorizontalFlip} and \textit{RandomAffine} for data augmentation during training.
Furthermore, during training, we applied \textit{RandomHorizontalFlip} and \textit{RandomAffine} transformations to the reference images as data augmentation.
\paragraph{Evaluation dataset.}
\begin{figure}[tb]
    \centering
    \includegraphics[width=1.0\linewidth]{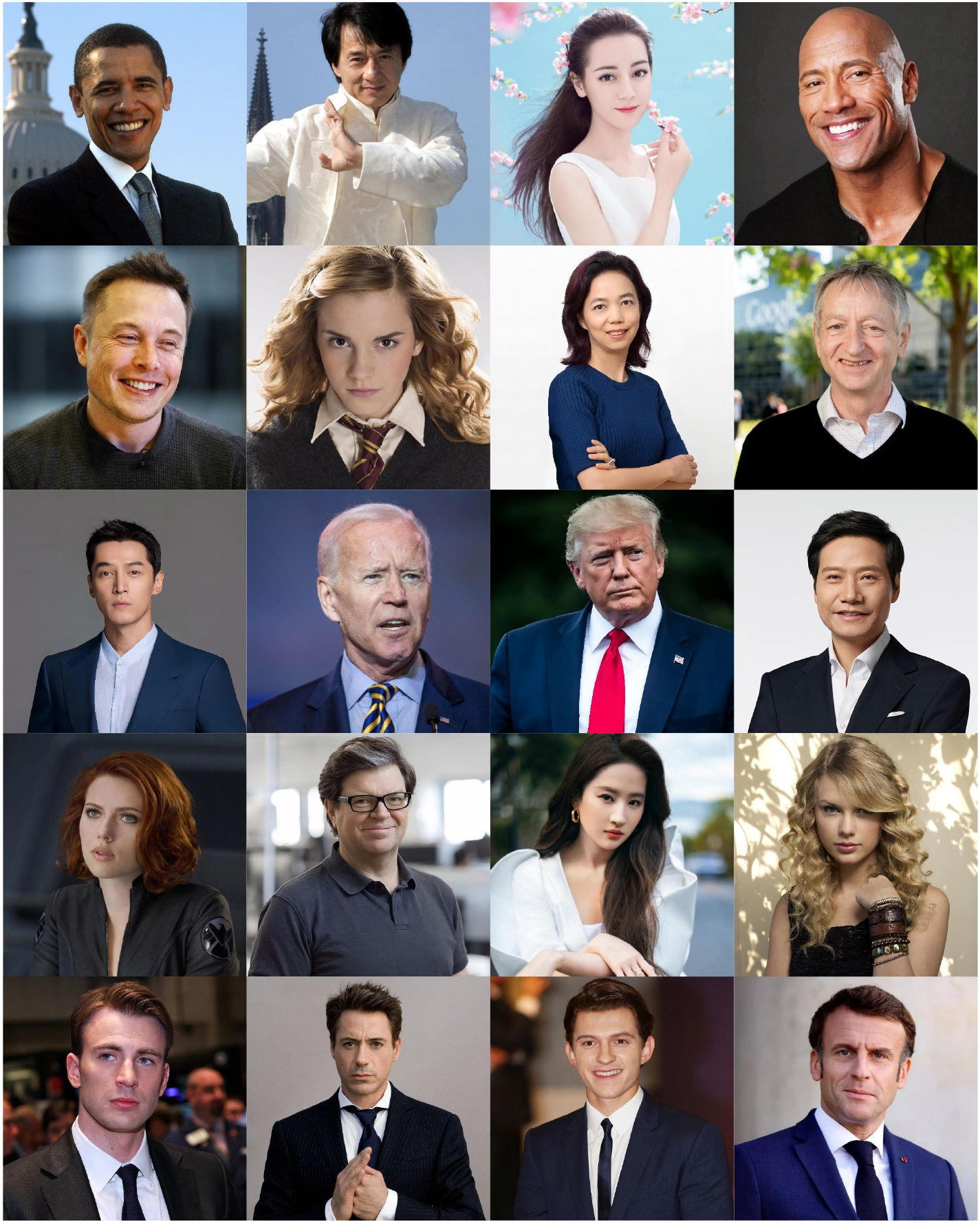}
    \caption{The overview of the celebrity dataset we use to test customized human video generation. }
    \label{fig:data_famous}
\end{figure}

\begin{figure}[tb]
    \centering
    \includegraphics[width=1.0\linewidth]{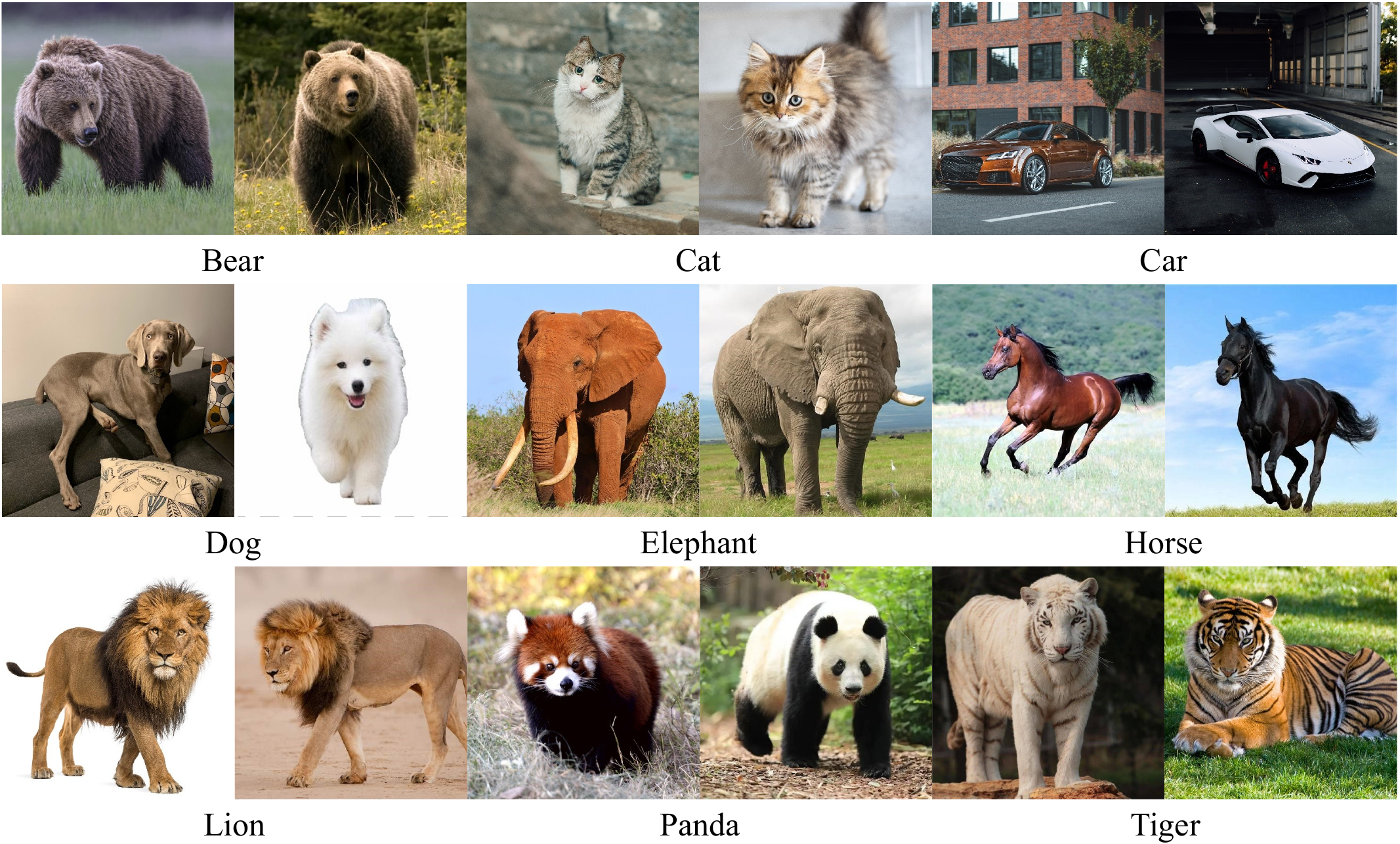}
    \caption{The overview of the dataset we use to test customized object video generation. }
    \label{fig:data_object}
\end{figure}

% \usepackage{multirow}
\begin{table}[]
    \centering
    \resizebox{\linewidth}{!}{
    \begin{tabular}{l|l}
    \toprule
    Category                    & Prompt                                                                                       \\
    \midrule
    \multirow{5}{*}{Clothing}   & A person dressed in a crisp white button-up shirt.                                           \\
                                & A person in a sleeveless workout top, displaying an active lifestyle.                        \\
                                & A person wearing a sequined top that sparkles under the light, ready for a festive occasion. \\
                                & A person wearing a Superman outfit.                                                          \\
                                & A person wearing a blue hoodie.                                                              \\
    \midrule
    \multirow{5}{*}{Action}     & A person holding a book open, reading a book, sitting on a park bench.                       \\
                                & A person playing an acoustic guitar.                                                         \\
                                & A person laughing with their head tilted back, eyes sparkling with mirth.                    \\
                                & A person is enjoying a cup of coffee in a cozy café.                                         \\
                                & A person watching a laptop, focused on the task at hand.                                     \\
    \midrule
    \multirow{5}{*}{Accessory}  & A person wearing a headphones, engaged in a hands-free conversation.                         \\
                                & A person with a pair of trendy headphones around their neck, a music lover's staple.         \\
                                & A person with a beanie hat and round-framed glasses, portraying a hipster look.              \\
                                & A person wearing sunglasses.                                                                 \\
                                & A person wearing a Christmas hat.                                                            \\
    \midrule
    \multirow{5}{*}{View}       & A person captured in a close-up, their eyes conveying a depth of emotion.                    \\
                                & A person framed against the sky, creating an open and airy feel.                             \\
                                & A person through a rain-streaked window, adding a layer of introspection.                    \\
                                & A person holding a bottle of red wine.                                                       \\
                                & A person riding a horse.                                                                     \\
    \midrule
    \multirow{5}{*}{Background} & A person is standing in front of the Eiffel Tower.                                           \\
                                & A person with a bustling urban street scene behind them, capturing the energy of the city.   \\
                                & A person standing before a backdrop of bookshelves, indicating a love for literature.        \\
                                & A person swimming in the pool                                                                \\
                                & A person stands in the falling snow scene at the park.                                      \\
    \bottomrule
    \end{tabular}
    }
    \caption{Evaluation text prompts for customized human video generation.}
    \label{tab:person_prompt}
\end{table}

% 我们在这里展示了我们正文Section 5.2所用的测试数据集。
% For customized human video generation, 我们follow了~\cite{li2024photomaker,wang2024instantid}等工作，收集了20个不同的people作为测试集，如图~\Cref{fig:data_famous}所示。
% 对于text prompt，我们consider five factors: clothing, accessories, actions, views, and background, which make up 25 prompts that are listed in the~\Cref{tab:person_prompt}进行测试。
% For customized object video generation, 由于VideoBooth~\cite{jiang2024videobooth}未公开其测试样本，我们从数据集中9个类别中每个类别收集了两个未出现在训练数据的样本进行测试。测试使用的prompt均根据object种类使用ChatGPT~\cite{achiam2023gpt}生成,具体prompt如~\Cref{tables/object_prompt}所示。
Here we present the test dataset used in \Cref{sc:cmp_value}.
For customized human video generation, we followed the works of \cite{li2024photomaker, wang2024instantid} and collected 20 different individuals as the test set, as shown in \Cref{fig:data_famous}. For the text prompts, we considered five factors: clothing, accessories, actions, views, and background, which make up 25 prompts listed in \Cref{tab:person_prompt} for testing.
% 在使用的时候，我们使用subject highlight preprocessing对参考图进行处理。
During inference, we processed the reference images using subject highlight preprocessing.
For customized object video generation, since VideoBooth~\cite{jiang2024videobooth} did not publicly release their test samples, we collected two samples from each of the nine categories that were not present in the training data for testing. The prompts used for testing were generated using ChatGPT~\cite{achiam2023gpt} based on the object categories, as detailed in \Cref{tab:object_prompt}.
During inference, we processed the reference images using subject highlight preprocessing and set the prompt for Grounding DINO~\cite{liu2023grounding} to "\texttt{<class word>.} " where \texttt{<class word>} represents the category of the object used, such as dog, cat.

\section{Quantitative Comparison Results on Non-Celebrity Dataset}
\begin{figure}[tb]
    \centering
    \includegraphics[width=1.0\linewidth]{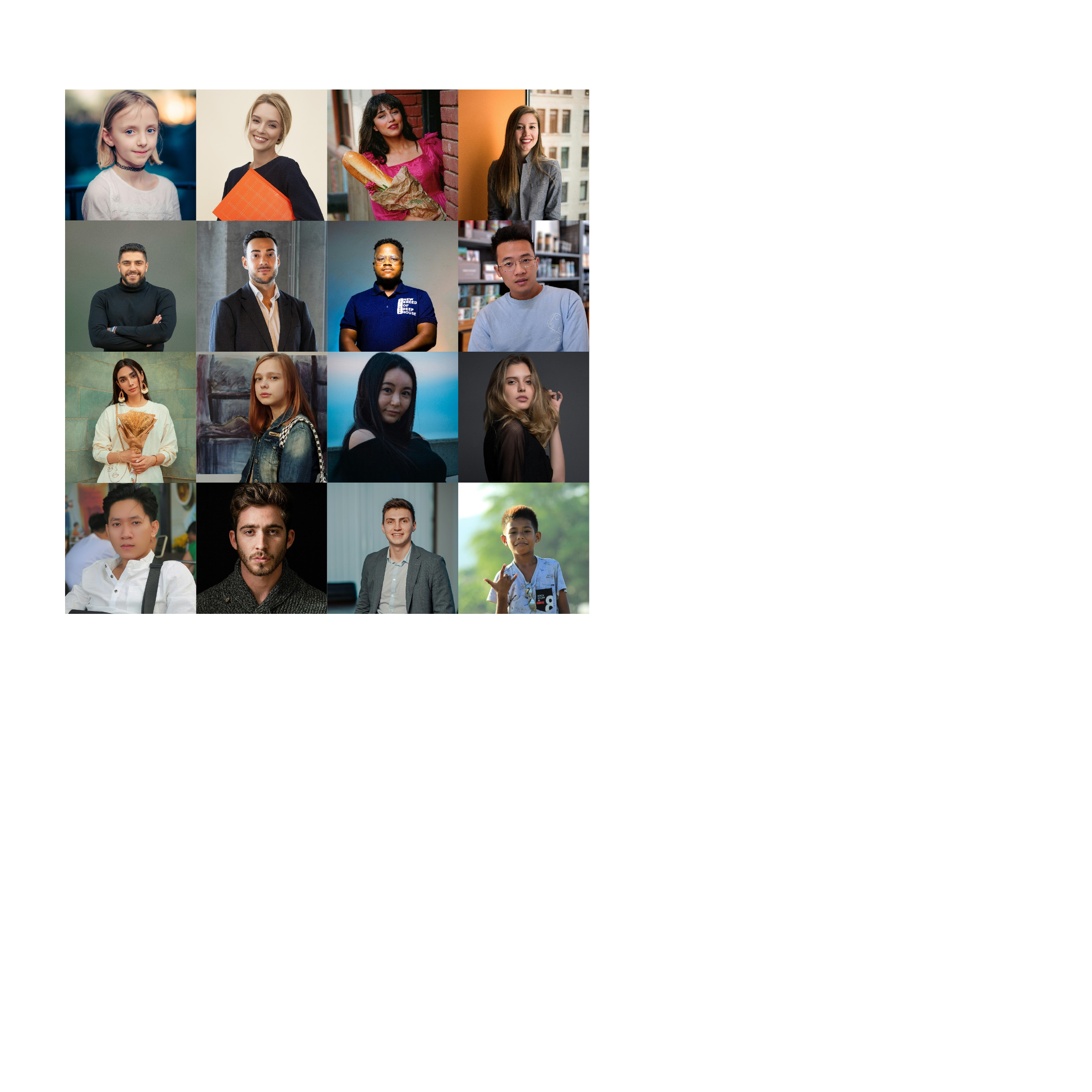}
    \caption{The overview of the non-celebrity dataset we used for testing customized human video generation.}
    \label{fig:data_common}
\end{figure}
% 一些研究~\cite{yuan2024inserting}指出由于预训练text-to-image diffusion model能够直接生成一些名人的照片,因此我们除了在测试时follow~\cite{li2024photomaker, wang2024instantid}等工作选取一些名人进行测试，也选取了一些非名人数据进行测试。
% 如图~\Cref{fig:data_common}所示,我们仿照~\cite{gal2024lcmlookahead}的Unsplash50数据集，从https://unsplash.com/收集了最新上传的a small set of 16 images with permissive licenses作为我们的非名人数据集，来确保这些图片从未出现在预训练数据中。
% 对于text prompt,我们采用了和名人一样的prompt的进行验证测试。
Some studies~\cite{yuan2024inserting} have pointed out that pre-trained text-to-image diffusion models can directly generate photos of certain celebrities.
Therefore, in addition to following works such as~\cite{li2024photomaker, wang2024instantid} by selecting some celebrities for testing, we also selected some non-celebrity data for testing.
As shown in ~\Cref{fig:data_common}, we followed the Unsplash50 dataset from~\cite{gal2024lcmlookahead} and collected a small set of 16 recently uploaded images with permissive licenses from https://unsplash.com/ as our non-celebrity dataset to ensure that these images have never appeared in the pre-training data. 
For the text prompts, we used the same prompts as those for celebrities.
% \usepackage[normalem]{ulem}
\begin{table}
    \centering
    \resizebox{\linewidth}{!}{
    \begin{tabular}{lcccccc}
    \toprule
    Method            & CLIP-T          & Face Sim.       & CLIP-I          & DINO-I          & T.Cons.         & DD              \\
    \midrule
    IP-Adapter        & 0.2347          & 0.1298          & 0.6364          & 0.5178          & \underline{0.9929}    & 0.0825          \\
    IP-Adapter-Plus   & 0.2140          & 0.2017          & \underline{0.6558}    & \underline{0.5488}    & 0.9920          & 0.0815          \\
    IP-Adapter-Faceid & 0.2457          & \underline{0.4651}    & 0.6401          & 0.4108          & 0.9930          & 0.0950          \\
    ID-Animator       & 0.2303          & 0.1294          & 0.4993          & 0.0947          & \textbf{0.9999} & 0.2645          \\
    Photomaker*       & \textbf{0.2803} & 0.2294          & 0.6558          & 0.3209          & 0.9768          & \underline{0.3335}    \\
    Ours              & \underline{0.2773}    & \textbf{0.6974} & \textbf{0.6882} & \textbf{0.5937} & 0.9797          & \textbf{0.3590} \\
    \bottomrule
    \end{tabular}
    }
    \caption{Comparison with the existing methods for customized human video generation on our non-celebrity dataset. The best and the second-best results are denoted in bold and underlined, respectively. Besides, PhotoMaker~\cite{li2024photomaker} is base on AnimateDiff~\cite{he2023animate} SDXL version.}
    \label{tab:common_sota}
\end{table}

% 定量比较结果如~\Cref{tab:common_sota}所示，我们的方法在非名人的数据集上依然展现了良好的性能。
% 非名人数据集上所有方法因为失去了一定的先验知识，均存在一定的指标略微下降，但是定量比较的结论与使用名人数据集的结论较为一致。
% 我们的方法依旧在衡量subject fidelity的Face Similarity, CLIP-I, DINO-I三个指标上明显领先。对于text alignment，我们的方法在使用AnimateDiff SD1.5 version作为基础模型的方法中取得取得了最好的效果。PhotoMaker的基础模型是AnimateDiff SDXL version，在基础模型层面拥有更强大的生成能力，但我们的方法能够取得较为接近的成果，这说明了我们采用模型原生能力进行subject信息注入的方式能在实现高保真的subject外观一致性的基础上，保证生成视频与给定prompt的匹配。
% 此外,我们的方法具有更好的动态度。

The quantitative comparison results are shown in~\Cref{tab:common_sota}. 
Our method still demonstrates good performance on the non-celebrity dataset. 
All methods show a slight decrease in metrics on the non-celebrity dataset due to the loss of certain prior knowledge, but the conclusions from the quantitative comparison are largely consistent with those using the celebrity dataset. 
Our method continues to lead significantly in the three metrics measuring subject fidelity: Face Similarity, CLIP-I, and DINO-I. 
For text alignment, our method achieves the best results among those using the AnimateDiff SD1.5 version as the base model. PhotoMaker uses the AnimateDiff SDXL version as its base model, which has a more powerful generative capability at the base model level. 
However, our method achieves comparable results, indicating that our approach of injecting subject information using the model's native capabilities can ensure high-fidelity subject appearance consistency while maintaining alignment between the generated video and the given prompt.
Additionally, our method exhibits better dynamism.
% 我们提供了非名人的定性比较结果可参见~\Cref{fig:cmp_non_cele,fig:cmp_non_cele2}.
\section{User Study}
\begin{figure}[tb]
    \centering
    \includegraphics[width=1.0\linewidth]{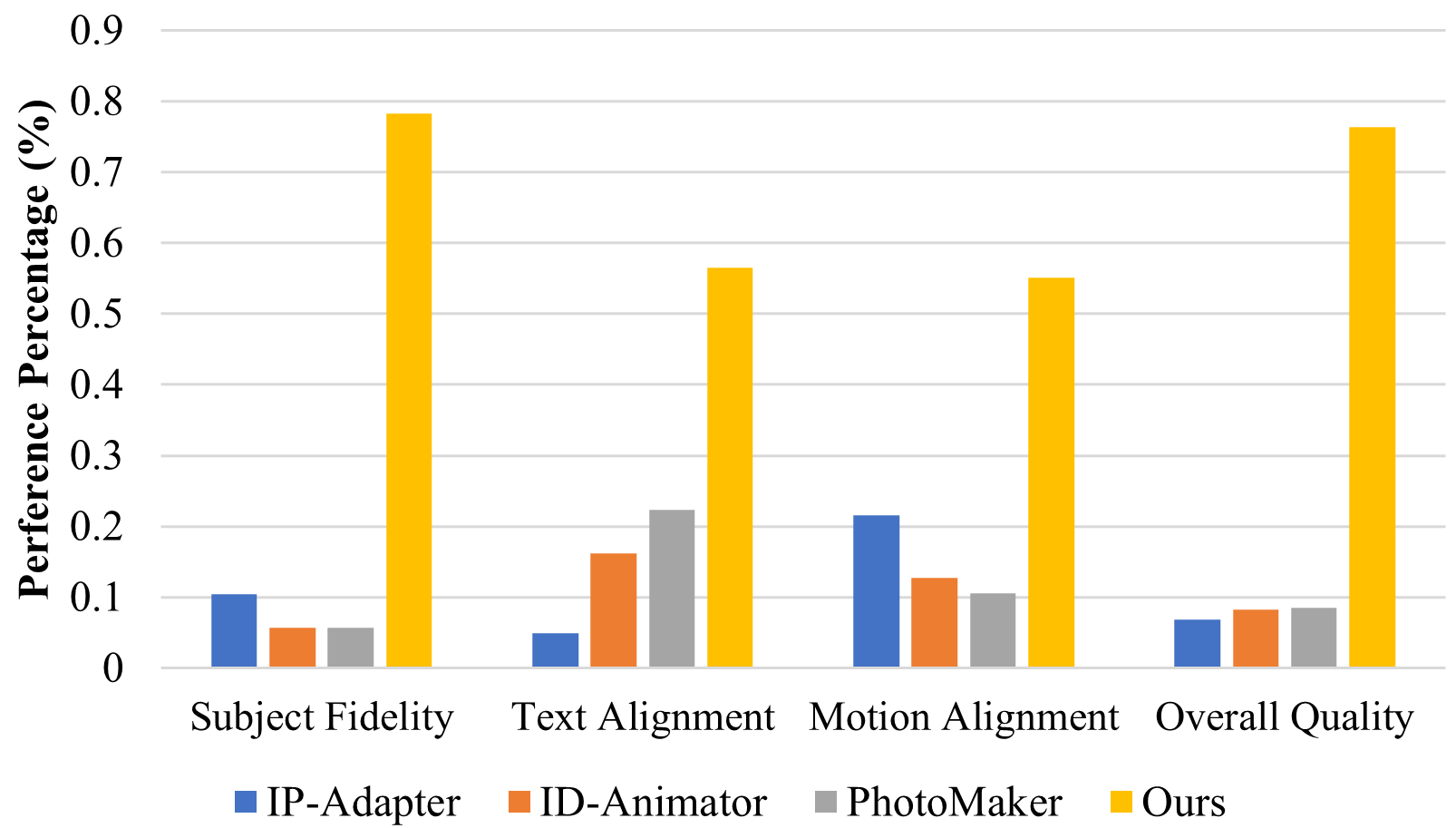}
    \caption{User Study for Customized Human Video Generation.}
    \label{fig:user_human}
\end{figure}

To further validate the effectiveness of our method, we conducted a human evaluation comparison of our method and existing methods. 
% We invited 5 professionals to evaluate the 30 sets of generated video results.
% For customized human video generation, 我们选用了10个名人和10个非名人作为测试基准。对于每个人，我们抽取两个prompt生成视频。我们邀请了10个专业人员对方法进行评测。
For customized human video generation, we selected 10 celebrities and 10 non-celebrities as the test benchmark. For each individual, we used two prompts to generate videos. We invited 10 professionals to evaluate the methods.
We evaluated the quality of the generated videos from four dimensions: Text Alignment, Subject Fidelity, Motion Alignment, and Overall Quality.
Text Alignment evaluates whether the generated video matches the text prompt. 
Subject Fidelity measures whether the generated object is close to the reference image. 
Motion Alignment is used to evaluate the quality of the motions in the generated video. 
Overall Quality is used to measure whether the quality of the generated video overall meets user expectations.
% 结果如~\Cref{fig:user_human}所示，我们的方法在各项评价指标上都取得了明显更多用户的喜爱，并在与subject fidelity有了显著提升，从而证明了我们方法的有效性。
As shown in \Cref{fig:user_human}, our method received significantly more user preference across various evaluation metrics. Additionally, it demonstrated a notable improvement in subject fidelity, thereby proving the effectiveness of our framework.
\begin{figure}[tb]
    \centering
    \includegraphics[width=1.0\linewidth]{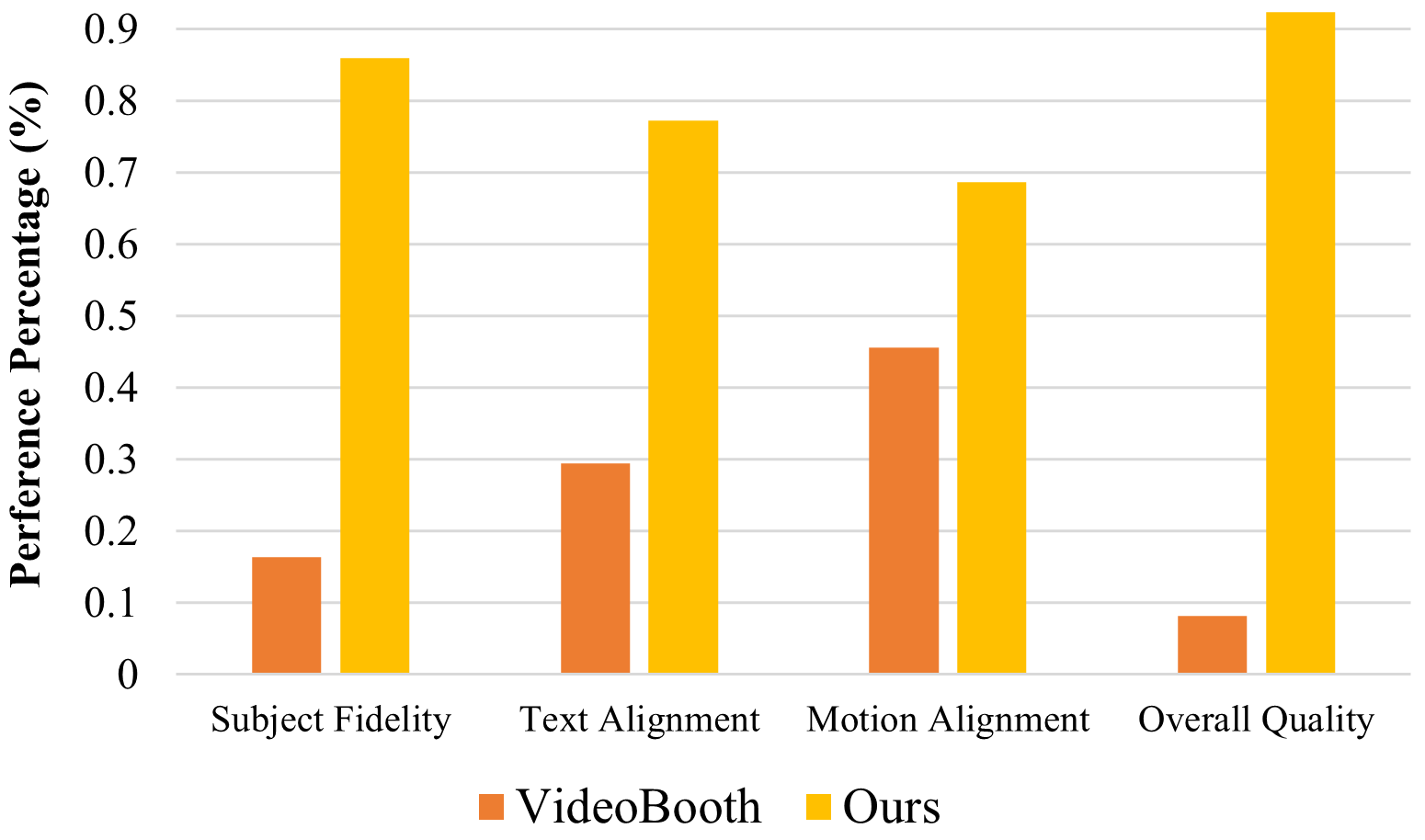}
    \caption{User Study for Customized Object Video Generation.}
    \label{fig:user_object}
\end{figure}
% For customized object video generation, 我们对VideoBooth数据集所包含的9个类别的物品都进行了主观评测。每个类别均提供一个subject,并给定2个ChatGPT生成的prompt进行测试。同样邀请了10个专业人员对方法进行评测。结果如~\Cref{fig:user_object}所示，相比于VideoBooth，我们方法在所有方面都得到了更多评测着的喜爱。

For customized object video generation, we conducted subjective evaluations on the 9 categories of objects included in the VideoBooth dataset. Each category provided one subject, and two prompts generated by ChatGPT~\cite{achiam2023gpt} were used for testing. We similarly invited 10 professionals to evaluate the methods. As shown in \Cref{fig:user_object}, our method received more favorable evaluations in all aspects compared to VideoBooth.

\section{Limitations and Future Work}
% 尽管我们的方法已经达到一个不错的高保真zero-shot customized video generation的效果，但是我们的方法仍然存在一些局限性。 
Our method only focuses on maintaining a single subject in the generated videos, and cannot control multiple subjects of generated persons in one video simultaneously.
In addition, our method, which is based on AnimateDiff and the dataset we utilized, inherits certain biases and limitations from these sources.

\paragraph{Limitations of the base model.} Our method is based on the SD1.5 version of AnimateDiff, and thus is limited by the generative capabilities of the base model. This can result in issues such as abnormal rendering of hands and limbs in the generated videos. Besides, since AnimateDiff inserts and fine-tunes Motion Blocks on the original image model, the base model's generated videos may exhibit poor dynamic effects, which in turn limits the dynamism of our method. Additionally, the base model has issues with facial clarity when the face is small in the generated images, affecting our customized portrait generation by failing to inject facial details well when the face occupies a smaller portion of the image. \
However, to ensure fair comparison with other methods and due to the limitations of our experimental equipment, we have not yet conducted experiments on better open-source models such as VideoCrafter~\cite{chen2023videocrafter1, chen2024videocrafter2}, CogVideoX~\cite{yang2024cogvideox}, and Latte~\cite{ma2024latte}. 
In the future, we will attempt to use more powerful base models to achieve better generative effects.

\paragraph{Limitations of the training datasets.} For customized human video generation: The CelebV-Text~\cite{yu2023celebv} dataset mainly consists of half-body videos, resulting in the model we trained on this dataset performing poorly in generating full-body videos. 
Our method excels at generating half-body portrait videos but is relatively less proficient at generating full-body portrait videos. Additionally, due to the coarse-grained captions in the training data, fine-grained control is not achievable.
For customized object video generation: The VideoBooth~\cite{jiang2024videobooth} dataset contains only a limited set of nine categories, so the model trained on this dataset cannot achieve truly universal generation of all objects. 
Furthermore, since the training videos for VideoBooth dataset are sampled from the WebVid~\cite{bain2021frozen} dataset, which contains watermarks, our customized object generation model trained on this dataset also results in generated videos with watermarks. 
In the future, we can attempt to train on better high-quality datasets to achieve truly universal zero-shot customized generation.
% \paragraph{Limitations of the basic model.} 我们的方法基于AnimateDiff 的SD1.5版本，因此会受限于底膜的生成能力，出现生成画面中存在手部、肢体等不正常问题。由于AnimateDif是通过在原有图像模型上插入并微调Motion Block，导致基础模型生成视频存在动态效果较差的问题，从而限制了我们方法的动态效果。此外，基础模型存在生成画面内人脸过小时存在面部不清晰问题，因此我们的定制化人像生成会受这一点的影响，在画面中人脸占比较小时候无法很好的注入面部细节。 但是为了公平的与其他方法进行比较，同时受限于我们的实验设备，我们暂时没有在诸如VideoCrafter~\cite{chen2023videocrafter1,chen2024videocrafter2},CogVideoX~\cite{yang2024cogvideox},Latte~\cite{ma2024latte}等更好的开源模型上进行实验，未来我们将尝试能力更强的基础模型以达到更好的生成效果。
% \paragraph{Limitations of the training datasets.}For customized human video generation：CelebV-Text~\cite{yu2023celebv}数据集中基本为half body 视频，因此我们方法生成全身视频表现较差。我们的方法擅长生成半身肖像视频，但在生成全身肖像视频方面相对不太擅长。此外，由于训练数据caption较为粗粒度，无法做到精细控制。对于customized object video generation：由于VideoBooth~\cite{jiang2024videobooth}数据集只有有限的十个类别，因此从该数据集训练出的模型无法做到真正意义的所有物品的通用生成。此外由于VideoBooth训练视频采样于WebVid~\cite{bain2021frozen}数据集，存在有水印，我们定制化物品生成模型在这个数据集上进行训练，因此导致我们生成视频有水印。未来我们可以尝试在更好的高质量数据集上进行训练，从而达到真正的通用zero-shot定制化生成。

\section{More Qualitative Comparison Results.}
\label{sc:more_vis}
% 为了进一步的证明我们方法的有效性，我们在这里补充了更多用于定性比较的可视化图。
% For customized human video generation, 我们首先补充了一部分名人的定制化生成结果。如图~\Cref{fig:cmp_cele}所示，我们的方法相比于现有的zero-shot定制化生成方法在保证text alignment的前提下展现出了更强的subject fidelity。我们方法生成的视频具有更多的面部细节。
% 例如~\Cref{fig:cmp_cele}(c), 我们的方法不仅相对于其他方法能够准确的刻画``enjoying a cup of coffe"的动作，并且在其他方法都未能很好保持subject外貌一致性的情况下，我们的方法取得了较高的subject fidelity。
% 此外，我们进一步展示了我们方法对于非名人的数据集的生成效果。如图~\Cref{fig:cmp_non_cele,fig:cmp_non_cele2}所示，我们方法在非名人数据上依旧可以实现高保真的zero-shot定制化生成.相比于现有方法有着更好的subject fidelity。
% 例如~\Cref{fig:cmp_cele}(f), 我们的方法可以准确的根据参考图和文本提示生成指定subject的视频，相比与其他方法展现出了明显的优势。

% For customized object video generation, 我们训练所使用的VideoBooth数据集有9个类别的物品，因此，我们补充了所有9类物品的定性比较。如图~\Cref{fig:cmp_object_appendxi}所示，我们的方法相比于VideoBooth在text alignment和subject fidelity都取得了明显的提升。
% 如~\Cref{fig:cmp_object_appendxi}(a,g)所示，我们的方法可以正确的生成'snowy'的正确场景，但是VideoBooth未能很好的生成对应场景。此外~\Cref{fig:cmp_object_appendxi}(i)中我们的方法正确生成了'a field of wildflowers'的场景，但是VideoBooth没有。
% 在subject fidelity方面，我们的方法相比于VideoBooth有着明显的提升，如~\Cref{fig:cmp_object_appendxi}(a,c,d,e,f,g,h,i)所示，对于这些动物，我们的方法在大场景下能够精确的刻画参考subject的纹理细节，而VideoBooth没有做到这一点。

\begin{figure*}[tb]
    \centering
    \includegraphics[width=1.0\linewidth]{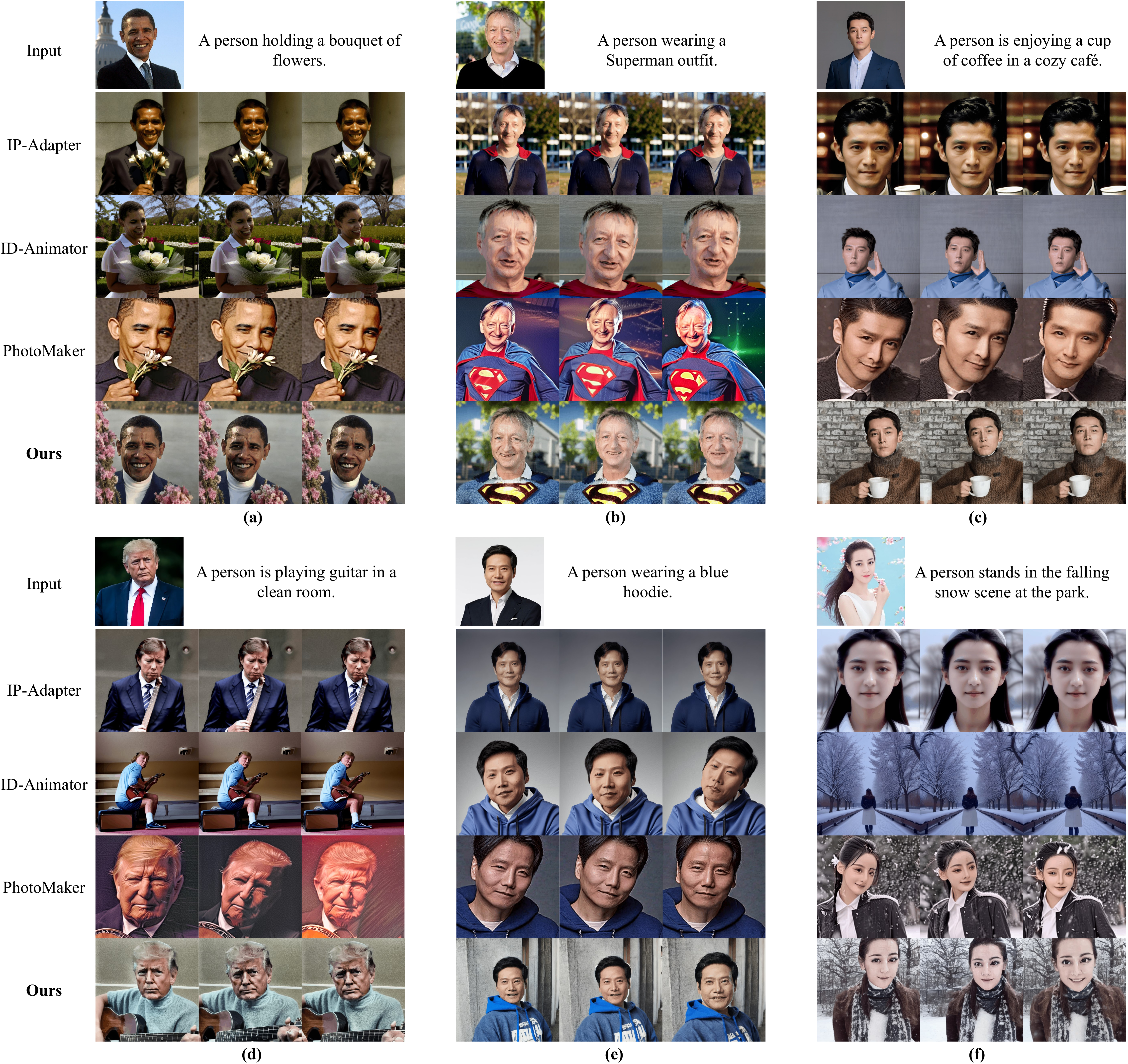}
    \caption{More Qualitative comparison for customized human video generation on celebrity dataset.}
    \label{fig:cmp_cele}
\end{figure*}

\begin{figure*}[tb]
    \centering
    \includegraphics[width=1.0\linewidth]{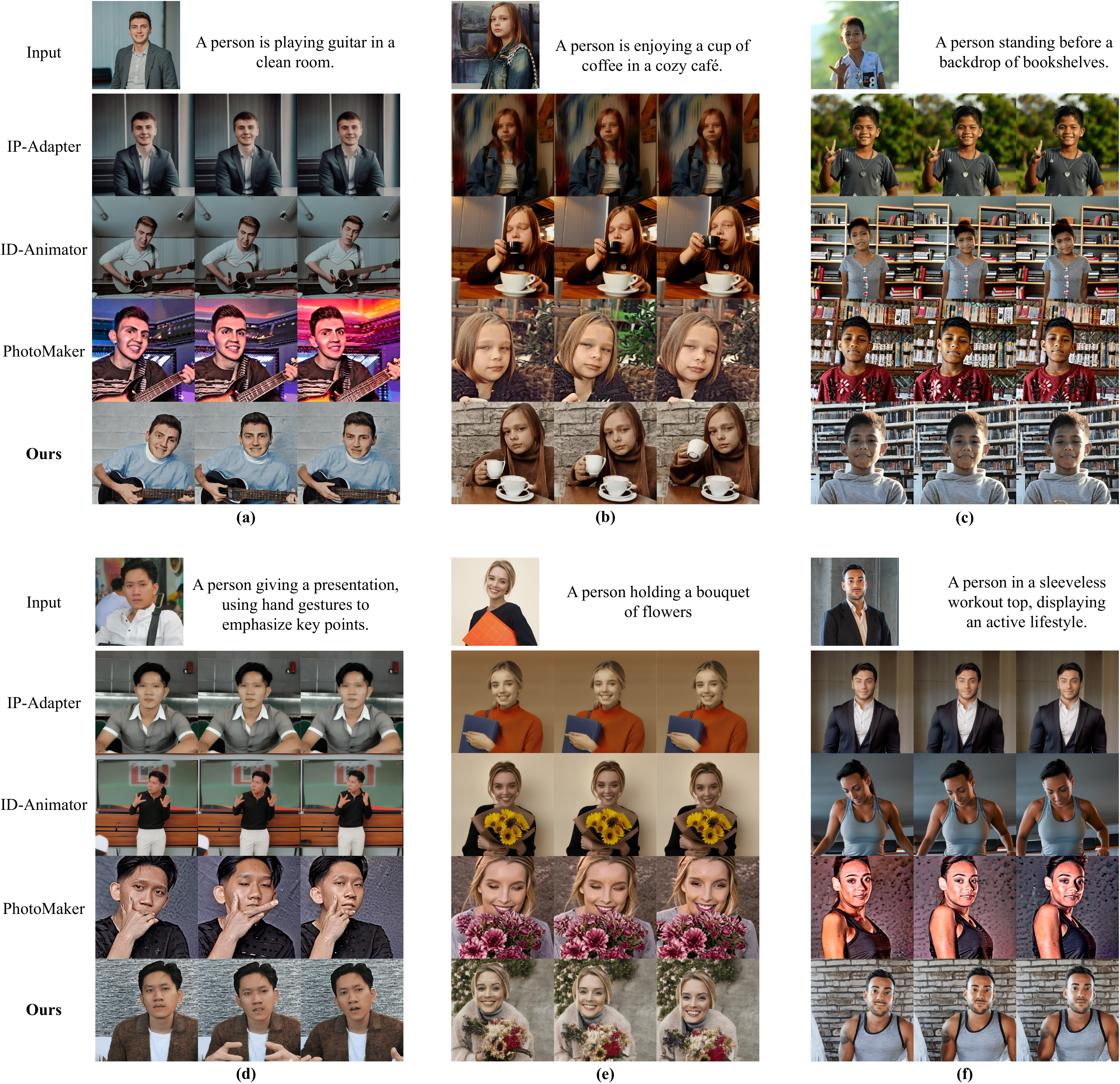}
    \caption{More Qualitative comparison for customized human video generation on non-celebrity dataset.}
    \label{fig:cmp_non_cele}
\end{figure*}

\begin{figure*}[tb]
    \centering
    \includegraphics[width=1.0\linewidth]{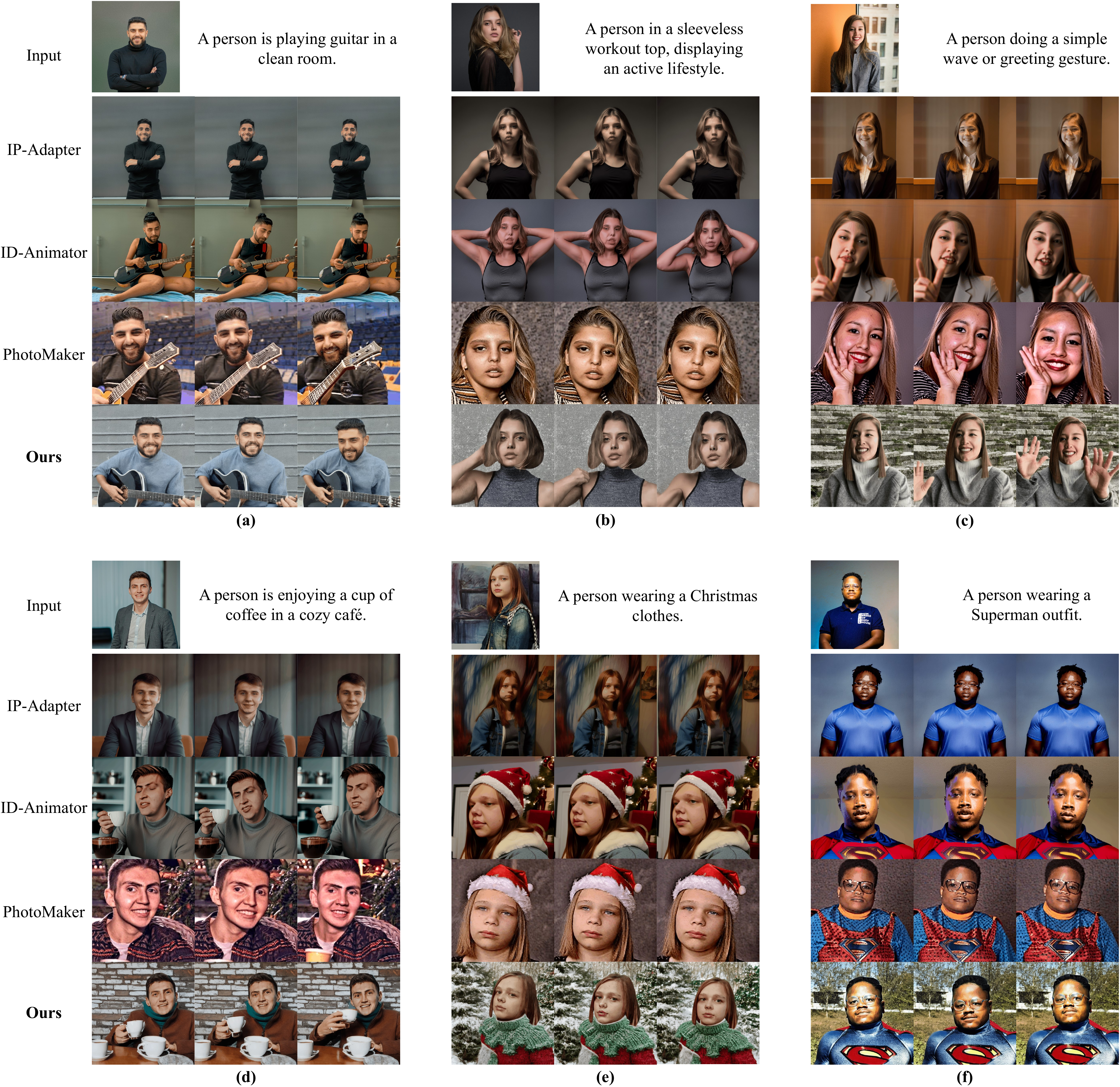}
    \caption{More Qualitative comparison for customized human video generation on non-celebrity dataset.}
    \label{fig:cmp_non_cele2}
\end{figure*}

To further demonstrate the effectiveness of our method, we have supplemented additional visualizations for qualitative comparison.
For customized human video generation, we first added some customized generation results for celebrities. As shown in \Cref{fig:cmp_cele}, our method exhibits stronger subject fidelity compared to existing zero-shot customization methods while ensuring text alignment. The videos generated by our method contain more facial details.
For example, in \Cref{fig:cmp_cele} (c), our method not only accurately depicts the action of "enjoying a cup of coffee" compared to other methods but also achieves high subject fidelity, maintaining the subject's appearance consistency where other methods fail to do so.
Additionally, we further demonstrate the generation effects of our method on the non-celebrity dataset. As shown in \Cref{fig:cmp_non_cele,fig:cmp_non_cele2}, our method can still achieve high-fidelity zero-shot customized generation on non-celebrity data, with better subject fidelity compared to existing methods.
For example, in \Cref{fig:cmp_cele} (f), our method accurately generates a video of the specified subject based on the reference image and text prompt, demonstrating a clear advantage over other methods.

\begin{figure*}[tb]
    \centering
    \includegraphics[width=1.0\linewidth]{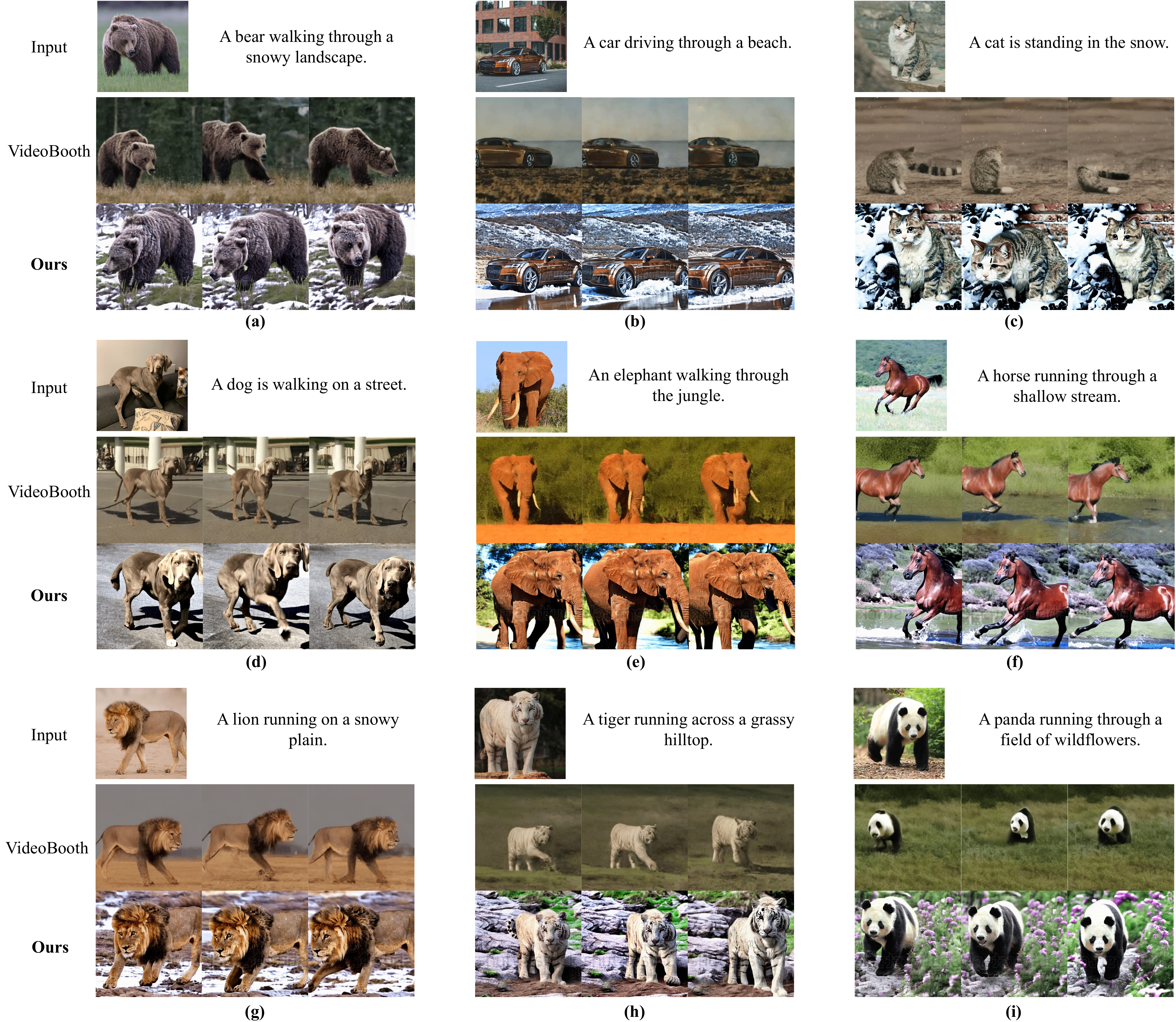}
    \caption{More Qualitative comparison for customized object video generation.}
    \label{fig:cmp_object_appendxi}
\end{figure*}

For customized object video generation, the VideoBooth dataset we used for training contains nine categories of objects. Therefore, we supplemented qualitative comparisons for all nine categories. As shown in \Cref{fig:cmp_object_appendxi}, our method achieves significant improvements in both text alignment and subject fidelity compared to VideoBooth.
As illustrated in \Cref{fig:cmp_object_appendxi} (a, g), our method correctly generates the 'snowy' scene, whereas VideoBooth fails to generate the corresponding scene accurately. Additionally, in \Cref{fig:cmp_object_appendxi} (i), our method correctly generates the scene of 'a field of wildflowers,' which VideoBooth does not.
In terms of subject fidelity, our method shows significant improvements over VideoBooth. As shown in \Cref{fig:cmp_object_appendxi} (a, c, d, e, f, g, h, i), for these animals, our method can accurately depict the texture details of the reference subject in large scenes, which VideoBooth fails to achieve. 
\section{Potential Societal Impacts}
In this paper, we present VideoMaker, a novel framework that leverages the inherent force of VDM to achieve zero-shot customized generation. Compared to heuristic external models for subject feature extraction and injection, we cleverly use VDM to accomplish the extraction and injection of subject features required for customized generation, resulting in high-quality customized video generation.

In practical applications, our method can be used in the film or video game industry to directly generate some required film clips through customized video generation. It can also be applied in virtual reality to provide a more immersive and personalized experience.

However, we acknowledge the ethical considerations that come with the ability to generate high-fidelity videos of humans or objects. 
The proliferation of this technology could lead to the misuse of generated videos, infringing on personal privacy rights, and potentially causing a surge in maliciously altered videos and the spread of false information. 
Therefore, we emphasize the importance of establishing and adhering to ethical guidelines and using this technology responsibly.
% \usepackage{multirow}
\begin{table*}[]
    \centering
    \resizebox{\linewidth}{!}{
    \begin{tabular}{l|ll|l}
    \toprule
    Category                   & Prompt                                                                              & Category                & Prompt                                                                    \\
    \midrule
    \multirow{10}{*}{bear}     & A bear walking through a snowy landscape.                                           & \multirow{10}{*}{car}   & A car cruising down a scenic coastal highway at sunset.                   \\
                               & A bear walking in a sunny meadow.                                                   &                         & A car silently gliding through a quiet residential area.                  \\
                               & A bear resting in the shade of a large tree.                                        &                         & A car smoothly merging onto a highway.                                    \\
                               & A bear walking along a beach.                                                       &                         & A car driving along a desert road.                                        \\
                               & A bear fishing in a rushing river.                                                  &                         & A car speeding through a muddy forest trail.                              \\
                               & A bear running in the forest.                                                       &                         & A car drifting around a sharp corner on a mountain road.                  \\
                               & A bear walking along a rocky shoreline.                                             &                         & A car navigating through a snow-covered road.                             \\
                               & A bear drinking from a clear mountain stream.                                       &                         & A car driving through a tunnel with bright lights.                        \\
                               & A bear standing on its hind legs to look around.                                    &                         & A car driving through a beach.                                            \\
                               & A bear running on the grass.                                                        &                         & A car driving through a foggy forest road.                                \\
    \midrule
    \multirow{10}{*}{cat}      & A cat is perched on a bookshelf, silently observing the room below.                 & \multirow{10}{*}{dog}   & A dog is lying on a fluffy rug, its tail curled neatly around its body.   \\
                               & A cat is sitting in a cardboard box, perfectly content in its makeshift fortress.   &                         & A dog is walking on a street.                                             \\
                               & A cat is curled up in a human's lap, purring softly as it enjoys being petted.      &                         & A dog is swimming.                                                        \\
                               & A cat is circle around a food bowl in a room, patiently waiting for mealtime.       &                         & A dog is sitting in a window, watching the raindrops race down the glass. \\
                               & A cat is lying on a windowsill, its silhouette framed by the setting sun.           &                         & A dog is running.                                                         \\
                               & A cat is running on the grass.                                                      &                         & A dog, a golden retriever, is seen bounding joyfully towards the camera.  \\
                               & A cat is walking on a street. There are many buildings on both sides of the street. &                         & A dog is seen leaping into a sparkling blue lake, creating a splash.      \\
                               & A cat is sitting in a window, watching the raindrops race down the glass.           &                         & A dog is seen in a snowy backyard.                                        \\
                               & A cat is playing with a ball of wool on a child bed.                                &                         & A dog is seen napping on a cozy rug.                                      \\
                               & A cat is playing in the snow, rolling and rolling, snowflakes flying.               &                         & A dog is seen playing tug-of-war with a rope toy against a small child.   \\
    \midrule
    \multirow{10}{*}{elephant} & An elephant walking through the jungle.                                             & \multirow{10}{*}{horse} & A horse walking through a dense forest.                                   \\
                               & An elephant crossing a river.                                                       &                         & A horse running across a grassy meadow.                                   \\
                               & An elephant walking on the grass.                                                   &                         & A horse walking along a sandy beach.                                      \\
                               & An elephant walking on a road.                                                      &                         & A horse running through a shallow stream.                                 \\
                               & An elephant walking along a dirt road.                                              &                         & A horse walking on a mountain trail.                                      \\
                               & An elephant playing in a mud pit.                                                   &                         & A horse running across a desert landscape.                                \\
                               & An elephant walking through a dense jungle.                                         &                         & A horse walking through a quiet village.                                  \\
                               & An elephant walking along a sandy beach.                                            &                         & A horse running in an open field.                                         \\
                               & An elephant running through a meadow of wildflowers.                                &                         & A horse walking along a forest path.                                      \\
                               & An elephant running across a desert landscape.                                      &                         & A horse running through tall grass.                                       \\
    \midrule
    \multirow{10}{*}{lion}     & A lion running along a savannah at dawn.                                            & \multirow{10}{*}{panda} & A panda walking through a bamboo forest.                                  \\
                               & A lion walking through a dense jungle.                                              &                         & A panda running on a grassy meadow.                                       \\
                               & A lion running on a snowy plain.                                                    &                         & A panda running through a field of wildflowers.                           \\
                               & A lion running along a rocky coastline.                                             &                         & A panda walking through a snowy landscape.                                \\
                               & A lion walking through a field of sunflowers.                                       &                         & A panda walking through a city park.                                      \\
                               & A lion running across a grassy hilltop.                                             &                         & A panda walking in front of the Eiffel Tower.                             \\
                               & A lion walking through a grassland.                                                 &                         & A panda wandering through a dense jungle.                                 \\
                               & A lion running along a riverbank.                                                   &                         & A panda running along a sandy beach.                                      \\
                               & A lion walking on a savannah during sunrise.                                        &                         & A panda exploring a cave.                                                 \\
                               & A lion running on a plain.                                                          &                         & A panda is eating bamboo.                                                 \\
    \midrule
    \multirow{5}{*}{tiger}     & A tiger running along a savannah at dawn.                                           & \multirow{5}{*}{tiger}  & A tiger running across a grassy hilltop.                                  \\
                               & A tiger walking through a dense jungle.                                             &                         & A tiger walking through a grassland.                                      \\
                               & A tiger running on a snowy plain.                                                   &                         & A tiger running along a riverbank.                                        \\
                               & A tiger running along a rocky coastline.                                            &                         & A tiger walking on a savannah during sunrise.                             \\
                               & A tiger walking through a field of sunflowers.                                      &                         & A tiger running on a plain.                                              \\
    \bottomrule
    \end{tabular}
    }
    \caption{Evaluation text prompts for customized object video generation.}
    \label{tab:object_prompt}
\end{table*}

\end{document}